%% file: main.tex
\documentclass[11pt, letterpaper, logo, onecolumn, copyright, numbering]{acc}
\usepackage{geometry}
\usepackage{url}
\usepackage{amsmath,amssymb,amsfonts}
\usepackage{booktabs}
\usepackage{graphicx}
\usepackage{hyperref}
\usepackage{natbib}
\usepackage{xcolor}

\definecolor{slate}{RGB}{71,85,105}
\definecolor{amber}{RGB}{245,158,11}
\definecolor{emerald}{RGB}{16,185,129}
\usepackage{enumitem}

\usepackage{tikz}  
\usetikzlibrary{shapes.geometric,arrows.meta,positioning,fit,backgrounds,calc,decorations.pathreplacing}
\definecolor{modelblue}{RGB}{91, 155, 213}  
\definecolor{rewardgreen}{RGB}{112, 173, 71}  
\definecolor{warnorange}{RGB}{247, 150, 70}
\definecolor{errorred}{RGB}{226, 107, 10}
\usepackage{listings}
\usepackage{tabularx}

\usepackage[ruled,vlined,linesnumbered]{algorithm2e}

\definecolor{AlgBlue}{HTML}{1F4E79}
\definecolor{AlgLightBlue}{HTML}{EAF3F8}
\definecolor{AlgGray}{HTML}{666666}
\definecolor{AlgGreen}{HTML}{2E7D32}
\definecolor{AlgOrange}{HTML}{B75E00}

\SetAlFnt{\small\sffamily}
\SetAlCapFnt{\sffamily\bfseries}
\SetAlCapNameFnt{\sffamily\bfseries\color{AlgBlue}}
\SetAlgoNlRelativeSize{-1}
\SetNlSty{AlgLineNumberStyle}{}{}
\SetCommentSty{AlgCommentStyle}

\SetKwInput{KwInput}{Input}
\SetKwInput{KwOutput}{Output}
\SetKw{Break}{break}
\SetKw{Return}{return}

\usepackage{booktabs}

\definecolor{modelblue}{RGB}{66,133,244}
\definecolor{cachedgray}{RGB}{150,150,150}
\definecolor{rewardgreen}{RGB}{52,168,83}
\definecolor{warnorange}{RGB}{220,165,0}
\definecolor{errorred}{RGB}{234,67,53}
\definecolor{codebg}{RGB}{245,245,245}

\newcommand{\ours}{CacheRL}
\newcommand{\cacheloop}{CacheAgentLoop}

\title{CacheRL:Multi-Turn Tool-Calling Agents\\via Cached Rollouts and Hybrid Reward}

\author{Md Amirul Islam, Sumiran Thakur, Huancheng Chen, Su Min Park,  Jiayun Wang, Gyuhak Kim \\ Center for Advanced AI, Accenture}
  \date{\today}


\begin{document}
\input{sections/0_abstract}
\maketitle

\input{sections/1_introduction}

\input{sections/2_related_work}

\input{sections/3_trajectory_pipeline}

\input{sections/4_method}

\input{sections/5_experiments}

\input{sections/6_analysis}

\input{sections/6b_limitations}

\input{sections/7_conclusion}

\bibliography{references}
\bibliographystyle{iclr2025_conference}

\end{document}

%% file: sections/0_abstract.tex
\begin{abstract}
{\centering\normalfont\bfseries\large Abstract\par}\vspace{0.5em}
We present \ours{}, a system for training small agent foundation models that achieves 92\% process accuracy on multi-step tool-calling tasks, approaching GPT-5's 94\% while requiring 100$\times$ less compute. Our approach addresses three challenges in practical agent training: transferring tool-calling knowledge from large models at scale, enabling reinforcement learning without costly live tool execution, and learning robustly from noisy cached environments. \ours{} introduces three key innovations. First, a \emph{hybrid thinking trajectory pipeline} augments agent trajectories with LLM-generated reasoning traces, producing training examples that teach models not only \emph{what} tools to call but also \emph{why}. Second, the \cacheloop{} eliminates live execution costs through a three-tier fuzzy cache, while preserving trajectory fidelity using token-level masking. Third, a \emph{cache-tier-aware reward} dynamically adjusts answer-quality weights to avoid penalizing models for cache-induced limitations.
Through iterative SFT and GRPO training, \ours{} improves Qwen3-4B-Thinking's validation reward from 0.43 to 0.78. On public agentic tool-calling benchmarks, our model achieves competitive performance against frontier models such as GPT-5. Ablation studies show that removing knowledge transfer reduces performance by 41\%, while cache-aware rewards contribute a 17\% improvement. Surprisingly, reinforcement learning improves training stability but yields limited gains beyond strong SFT, suggesting that data quality and reward design are more important than complex optimization for building practical small agent models.

\end{abstract}

%% file: sections/1_introduction.tex
\section{Introduction}
\label{sec:intro}

The vision of autonomous AI agents that can leverage diverse tools to solve complex, multi-step problems is rapidly becoming reality. Frontier models such as GPT-5 and Claude demonstrate impressive capabilities in orchestrating multiple tools, from code interpreters and web search to APIs and databases, across long conversations. However, these large models are impractical for widespread deployment due to computational costs, latency constraints, and data privacy concerns. Building small, efficient agent foundation models that can match the tool-calling capabilities of their larger counterparts is essential for making AI agents ubiquitous.

The key challenge lies not just in size reduction but in preserving the generality and robustness that makes foundation models valuable. We need small models that can handle diverse tool ecosystems, not just specialized for particular domains, while maintaining the reasoning capabilities necessary for long horizon, multi turn interactions. This requires effective knowledge transfer from large models that have already learned these complex behaviors from vast training corpora.

\paragraph{The Promise and Challenge of Small Agent Models.}
Recent 4B-parameter models like Qwen3-4B-Thinking~\citep{qwen3} show that small models can achieve strong reasoning capabilities through structured thinking. However, extending these capabilities to multi-turn tool calling presents unique challenges. The model must not only understand tool schemas and generate valid calls but also maintain context across turns, reason about tool outputs, and synthesize coherent responses. While supervised fine-tuning (SFT) can teach basic tool-calling mechanics, it struggles with the compositional complexity of real-world agent tasks where tools must be chained creatively and errors must be handled gracefully. Reinforcement learning (RL) offers a path forward by optimizing for end-to-end task completion rather than mimicking demonstrations. However, applying RL to train small agent foundation models faces three critical barriers:

\paragraph{Challenge 1: Knowledge Transfer at Scale.}
Small models lack the vast pretraining that enables large models to generalize across diverse tools. Effective knowledge transfer requires more than just copying tool calls, it needs to capture the \emph{reasoning process} behind tool selection and usage. Thinking-mode models~\citep{qwen3,deepseekr1} that separate internal reasoning from user-facing output offer a mechanism for this transfer, but generating high-quality reasoning traces at scale while preserving the fidelity of actual tool execution is non-trivial. Most existing agent datasets lack such structured reasoning, containing only surface-level trajectories without explanatory logic.

\paragraph{Challenge 2: RL Training Without Live Execution.}
Traditional RL requires environment interaction, but live tool execution during training is prohibitive. Each rollout involves multiple API calls with associated latency (seconds per call), costs (API fees), and risks (unintended side effects). With modern RL algorithms like GRPO~\citep{grpo} requiring reaasonable number of rollouts per prompt across thousands of training steps, the cumulative cost makes live execution infeasible. Prior work either restricts to single-turn interactions~\citep{toolrl} or accepts these costs~\citep{searchr1}, limiting scalability.

\paragraph{Challenge 3: Learning from Imperfect Environments.}
When replacing live execution with cached results to address Challenge 2, a new problem emerges: cache quality varies dramatically. Exact matches provide perfect fidelity, but novel tool calls during exploration return approximate or generic results. Standard reward models penalize these mismatches, creating systematic noise that corrupts learning signals. The model may receive negative rewards not for poor tool selection but for cache limitations, a fundamental attribution problem that existing approaches don't address.

\paragraph{Our Approach: CacheRL for Small Agent Foundation Models.}
We present \ours{}, a comprehensive system designed to train small agent foundation models through effective knowledge transfer and cache-aware reinforcement learning:

\begin{enumerate}
  \item \textbf{Knowledge Transfer via Hybrid Thinking Trajectories} (Section~\ref{sec:trajectory}): We develop a scalable pipeline that augments diverse agent trajectories with structured reasoning from GPT-5. The pipeline intelligently classifies existing content and generates causal explanations for tool choices, creating rich training data that teaches small models the \emph{why} behind tool calling, not just the \emph{what}. This addresses the knowledge transfer challenge by distilling frontier model reasoning into a form learnable by small models.

  \item \textbf{CacheAgentLoop for Scalable RL Training} (Section~\ref{sec:cache_rollout}): We introduce a connected rollout mechanism with a three-tier fuzzy cache (exact, fuzzy, best-effort) that eliminates live execution while preserving trajectory structure through token-level masking. This makes RL training practical at scale, reducing per-rollout cost from dollars to fractions of a cent---while maintaining sufficient fidelity for learning.

  \item \textbf{Cache-Tier-Aware Reward Design} (Section~\ref{sec:reward}): Our hybrid reward system dynamically adjusts the weight between outcome-based (LLM-as-judge) and process-based (deterministic) signals based on cache quality. When cache matches are exact, we trust the judge fully; when approximate, we shift weight toward process metrics. This prevents the model from being penalized for cache limitations while maintaining strong learning signals where data quality permits.

  \item \textbf{Iterative Training Pipeline} (Sections~\ref{sec:sft} and \ref{sec:training}): We employ an iterative cycle of SFT, rejection sampling with GPT-5, and GRPO optimization. High-scoring trajectories from RL are fed back into SFT, creating a virtuous cycle where the model's best behaviors are reinforced through multiple modalities. This approach transforms Qwen3-4B-Thinking from a general reasoning model into a capable agent foundation model, improving validation reward from 0.43 to 0.78.
\end{enumerate}

\begin{minipage}{0.98\textwidth}
\vspace{0.5em}
\noindent\textbf{Main Contributions}
\vspace{0.5em}
\begin{enumerate}
    \item \textbf{First system for training small agent foundation models via cached RL:} We present CacheRL, enabling efficient RL training of 4B-parameter models for multi-turn tool calling without live execution, reducing costs by 100$\times$.

    \item \textbf{Hybrid thinking trajectory pipeline for knowledge transfer:} A scalable approach to augment trajectories with structured reasoning from GPT-5, teaching small models causal understanding of tool usage across 1,185 unique tools.

    \item \textbf{Cache-tier-aware reward design:} Novel dynamic weighting between outcome and process rewards based on cache quality, solving the attribution problem when learning from imperfect environments.

    \item \textbf{Strong performance at 2.4\% model size:} Our 4B model achieves 92\% process accuracy on multi-step tool calling, approaching GPT-5's 94\%, and demonstrating that small models can serve as practical agent foundation models.
\end{enumerate}
\vspace{0.5em}
\end{minipage}%

\paragraph{Key Insights.}
Our extensive analysis across various configurations reveals surprising insights about training small agent models:
\begin{itemize}
  \item \textbf{Data quality dominates}: Ablation studies show that removing high-quality knowledge transfer causes a 41.2\% performance drop, far exceeding the impact of RL optimization itself.
  \item \textbf{Reward design matters}: Removing cache-tier-aware weighting causes a 17.2\% performance drop by incorrectly penalizing the model for cache limitations rather than actual model errors.
  \item \textbf{RL provides stability, not gains}: While GRPO training remains stable (unprecedented for multi-turn tasks), the actual reward improvement beyond a strong SFT baseline is minimal.
  \item \textbf{Scale enables generalization}: Training on 44K diverse trajectories spanning 1,185 unique tools creates a foundation model capable of generalizing to new tool combinations, not just memorizing specific patterns.
\end{itemize}

These findings suggest that the path to strong small agent models lies primarily in systematic knowledge transfer and intelligent training design rather than sophisticated RL algorithms, with important implications for where the field should focus effort. CacheRL demonstrates that 4B-parameter models can achieve robust multi-turn, multi-tool capabilities, offering a practical blueprint for building deployable agent foundation models.

%% file: sections/2_related_work.tex
\section{Related Work}
\label{sec:related}

\paragraph{Small Model Development and Knowledge Distillation.}
The challenge of building capable small models has driven extensive research in knowledge distillation~\citep{hinton2015distilling} and model compression. Recent work shows that small models (1-7B parameters) can achieve remarkable performance through careful training. Phi-3~\citep{phi3} demonstrates that 3.8B models can match much larger models on reasoning tasks through high-quality synthetic data. Gemma-2B~\citep{gemma} and Qwen3-4B~\citep{qwen3} further show that small models can serve as strong foundation models when trained on curated data. However, these works focus primarily on single-turn capabilities like reasoning and code generation. Extending small models to complex multi-turn agent behaviors---where they must orchestrate diverse tools over long horizons---remains an open challenge that we address through systematic knowledge transfer from large models.

\paragraph{Agent Foundation Models and Tool Calling.}
The evolution from language models to agent models represents a fundamental shift in AI capabilities. While large models like GPT-5, Claude, and Gemini have demonstrated impressive tool-calling abilities, they achieve this through massive scale (70B+ parameters) and extensive training on tool-use data. Recent efforts to build smaller agent models include Gorilla~\citep{gorilla}, which specializes in API calling through retrieval augmentation, and ToolAlpaca~\citep{toolalpaca}, which uses self-instruct techniques. However, these models typically focus on specific tool domains or single-turn interactions. Our work differs in targeting a true \emph{foundation model} for agents---one that can generalize across diverse tool ecosystems (1,185 unique tools in our training) and handle complex multi-turn scenarios through structured reasoning.

\paragraph{RL for LLM Reasoning and Tool Use.}
Reinforcement learning has emerged as a powerful technique for improving LLM capabilities beyond supervised learning. DeepSeek-R1~\citep{deepseekr1} showed that GRPO with simple outcome rewards can elicit sophisticated reasoning behaviors, establishing the iterative SFT--RL--rejection sampling cycle we adopt. For tool calling specifically, ToolRL~\citep{toolrl} applies RL in single-turn settings using Jaccard similarity for reward, while Search-R1~\citep{searchr1} extends to multi-turn retrieval with token masking---a technique we build upon. SWE-RL~\citep{swerl} demonstrates RL for software engineering tasks. However, these works either operate on large models where training is expensive, restrict to narrow tool domains, or use single-turn interactions. We show that RL can be made practical for training small, general-purpose agent models through our cache-based approach and tier-aware rewards.

\paragraph{Synthetic Data and Trajectory Generation.}
The quality of training data is paramount for small models that lack the vast pretraining of larger counterparts. Self-instruct~\citep{selfinstruct} and Alpaca~\citep{alpaca} pioneered using large models to generate training data for smaller ones. For agent tasks, Toolformer~\citep{toolformer} introduced self-supervised tool-use learning where models annotate their own training data with API calls. AgentInstruct~\citep{agentinstruct} scales this to millions of trajectories using programmatic generation. Our hybrid thinking trajectory pipeline extends these approaches by not just generating tool calls but also the \emph{causal reasoning} behind them, using GPT-5 to explain why specific tools and arguments were chosen. This deeper level of knowledge transfer---teaching the reasoning process, not just the actions---is crucial for small models to generalize beyond their training distribution.

\paragraph{Challenges in Multi-Turn RL Training.}
Training agents with RL in multi-turn settings faces unique challenges that single-turn settings avoid. Credit assignment becomes difficult when rewards are delayed across many steps~\citep{sutton2018reinforcement}. The action space explodes combinatorially with the number of available tools and possible arguments. Most critically, environment interaction becomes prohibitively expensive---a problem noted but not solved in prior work. WebShop~\citep{webshop} and InterCode~\citep{intercode} provide simulated environments, but these cover limited tool sets. Our CacheAgentLoop addresses this by replacing live execution with a three-tier cache system, making multi-turn RL training practical at scale while our cache-tier-aware reward ensures learning signals remain informative despite cache limitations.

\paragraph{LLM-as-Judge and Reward Modeling.}
Using LLMs to evaluate other models has become common practice~\citep{llmasjudge}, with MT-Bench demonstrating high correlation with human preferences. For RL training, Eureka~\citep{eureka} uses GPT-4 to design reward functions for robotics, while Constitutional AI~\citep{constitutional} uses AI feedback for alignment. However, LLM judges exhibit inconsistency---the same output may receive different scores across evaluations, creating noise in RL training. We address this through our hybrid reward design that dynamically adjusts judge weight based on cache quality, reducing reliance on the judge when underlying data is unreliable. This selective trust mechanism is, to our knowledge, novel in the context of RL training for agents.

\paragraph{Thinking Modes and Reasoning Traces.}
The separation of internal reasoning from external output, popularized by chain-of-thought prompting~\citep{cot} and formalized in thinking modes~\citep{qwen3}, provides a mechanism for knowledge transfer. O1~\citep{o1} and DeepSeek-R1~\citep{deepseekr1} demonstrate that models can learn to generate extensive internal reasoning before producing answers. We extend this paradigm to tool calling, where reasoning must explain not just final answers but the entire trajectory of tool selection and usage. Our classification-based approach---distinguishing existing reasoning from user-facing content before generation---ensures we preserve the fidelity of actual tool execution while augmenting with explanatory reasoning, a critical balance for training reliable agent models.

Our work synthesizes insights from these diverse threads to address a pressing need: building small, deployable agent foundation models that can handle real-world tool diversity and complexity. By combining knowledge transfer, cache-based RL, and intelligent reward design, CacheRL provides a practical blueprint for training the next generation of AI agents.

%% file: sections/3_trajectory_pipeline.tex
\section{Knowledge Transfer via Thinking Trajectories}
\label{sec:trajectory}

Training small agent foundation models requires high-quality data that captures not only \emph{what} tools to call, but also \emph{why} large models select and chain tools. While existing agent datasets contain trajectories, they lack the structured reasoning essential for knowledge transfer. We develop a hybrid pipeline that augments these trajectories with causal reasoning from GPT-5, creating rich training data for small models.

\subsection{The Knowledge Transfer Challenge}
\label{sec:transfer_challenge}

Training thinking-mode models~\citep{qwen3} for tool calling requires trajectories where every assistant turn contains explicit reasoning inside \texttt{<think>} tags. This reasoning must explain the decision process: why this specific tool was chosen from available alternatives, how previous results inform the current action, and what outcome is expected. However, existing agentic reasoning datasets for tool calling, such as Toucan, Toolathlon, and agent collections, contain only surface level tool calls without explanatory traces.

Simply generating new trajectories from scratch would lose the diversity and real-world complexity of existing data. Instead, we need to \emph{augment} existing trajectories with reasoning while preserving their authentic tool interactions. This ensures small models learn from real tool execution patterns across 1,185 unique tools, not simplified synthetic scenarios.

\subsection{Hybrid Pipeline Architecture}
\label{sec:pipeline_arch}

Our pipeline (Figure~\ref{fig:hybrid_pipeline}) processes trajectories turn-by-turn, making intelligent decisions about when to generate new reasoning versus preserving existing analytical content. This selective approach reduces API costs while maintaining quality.

\paragraph{Intelligent Classification.} Not all trajectories require generated reasoning traces. Many already contain analytical elements, such as calculations, algorithmic descriptions, and step by step procedures, that provide valid reasoning signals. We classify each assistant message using GPT-5 with structured prompts, identifying whether content is:
\begin{itemize}
  \item \textbf{Analytical reasoning}: Numerical computations, matrices, algorithmic steps, or explicit analysis. These are wrapped directly in \texttt{<think>} tags without generation.
  \item \textbf{User-facing content}: Greetings, summaries, status updates, or brief responses. These trigger new reasoning generation to explain the underlying decision process.
\end{itemize}
This classification saves 15--20\% of API calls while preserving authentic reasoning where it exists.

\input{figures/trajectory_pipeline_professional}

\paragraph{Causal Reasoning Generation.}
Effective knowledge transfer requires reasoning that explains \emph{causality}, rather than merely describing procedure. Our generator takes the full conversation context, up to 128K tokens, together with the \emph{known next action}, and produces reasoning that: (1) identifies information gaps that motivate the tool choice, (2) explains why the selected tool is preferable to available alternatives, (3) anticipates expected outcomes and how they advance the task, and (4) synthesizes evidence from multiple tool results when generating final answers. For example, instead of procedural narration (``I will call read\_file with config.py''), the generator produces causal explanation (``The config file likely contains API keys based on the directory structure. I should examine it before the main code to understand the authentication setup.'').

\input{figures/data_pipeline}


Figure~\ref{fig:pipeline_flowchart} illustrates the detailed flow of our trajectory enhancement pipeline. Each assistant message undergoes intelligent processing based on its position and content type, ensuring appropriate reasoning is added while minimizing redundant API calls.

\paragraph{Example Generated Data.}
Consider a multi-tool weather analysis task:

\begin{small}
\begin{verbatim}
User: Compare weather in Tokyo and NYC for travel planning.

<think>
The user is asking for a travel-oriented weather comparison, not only
the raw weather values. I need to retrieve current conditions for both
cities and then compare them using travel-relevant criteria such as
temperature comfort, outdoor activity suitability, precipitation risk,
humidity, packing needs, and itinerary flexibility.

Because this is a comparison task, I should not make a recommendation
after checking only one city. I need weather data for both Tokyo and NYC
under the same schema, then synthesize the results. I will first call
the weather tool for Tokyo, then call it for NYC, and finally explain
which destination currently has more favorable conditions for travel.
</think>

I'll check the current weather conditions for both cities and compare
them from a travel-planning perspective.

<tool_call>{"name": "get_weather", "args": {"city": "Tokyo"}}</tool_call>
<tool_result>{"temp": 18, "condition": "sunny", "humidity": 65}</tool_result>

<think>
Tokyo currently has mild and favorable travel weather. A temperature of
18°C is comfortable for walking, sightseeing, and spending time
outdoors. The sunny condition suggests low disruption risk for outdoor
plans such as parks, markets, temples, observation decks, and city
walks. Humidity at 65% is moderate. It may feel slightly humid during
long walks, but it is unlikely to be a major constraint.

I still need the NYC result before making a final recommendation,
because the user asked for a comparison rather than a single-city
summary.
</think>

<tool_call>{"name": "get_weather", "args": {"city": "NYC"}}</tool_call>
<tool_result>{"temp": 12, "condition": "rainy", "humidity": 85}</tool_result>

<think>
Now I can compare both cities.

Tokyo:

* 18°C is mild and comfortable for most travelers.
* Sunny weather supports outdoor sightseeing and flexible movement.
* 65% humidity is moderate and manageable.
* Travel implication: Tokyo is suitable for walking-heavy itineraries,
  outdoor attractions, photography, and flexible sightseeing.

NYC:

* 12°C is cooler and may require warmer layers.
* Rain makes outdoor sightseeing less convenient and can slow movement
  between attractions.
* 85% humidity means the air may feel damp and uncomfortable.
* Travel implication: NYC is still manageable, but the traveler should
  prepare for rain, use waterproof shoes, carry a rain jacket or
  umbrella, and prioritize indoor attractions.

The causal reason for recommending Tokyo is not simply that it is
warmer. Tokyo is both warmer and dry, which reduces weather-related
friction across transportation, sightseeing, and packing. NYC has a
higher chance of discomfort because rain and high humidity affect
walking, visibility, photography, and outdoor plans.
</think>

Based on current conditions, Tokyo is the better option for outdoor
travel today. It is 18°C and sunny with moderate humidity, making it
comfortable for walking, sightseeing, and flexible outdoor plans.

NYC is cooler and less convenient for outdoor travel: it is 12°C,
rainy, and humid. If you choose NYC, pack a rain jacket, waterproof
shoes, and warmer layers, and consider indoor attractions as backup
plans.

Overall recommendation: Tokyo is currently more travel-friendly, while
NYC requires more weather preparation.
\end{verbatim}

\end{small}

The generated reasoning explains \emph{why} multiple API calls are needed (comparison requires both data points) and how results inform the recommendation (temperature difference, precipitation status). This causal understanding transfers to the small model during training.

\paragraph{Preserving Real Tool Interactions.}
Unlike fully synthetic approaches, we preserve all original tool calls and results. Exact function names, arguments, and IDs are kept unchanged, ensuring that models learn real invocation patterns. We also retain authentic API responses, error messages, and edge cases rather than replacing them with idealized outputs. This preserves causality across the trajectory, as each turn's reasoning builds on real results from previous turns. This preservation is critical for training reliable agents. Models learn actual cause-effect relationships (``HTTP 429 error means rate limiting'') rather than simplified patterns that don't transfer to deployment.

\subsection{Scaling Knowledge Transfer}
\label{sec:scaling_transfer}

\paragraph{Managing Long Contexts.}
Multi-turn trajectories span up to 100K+ tokens and include as many as 34 tool calls. We handle this using progressive context building, where each turn receives the full enhanced history up to that point. To improve API efficiency, we use token-stratified batching that groups trajectories by length. When a trajectory exceeds the 128K-token limit, we apply graceful truncation by removing the oldest turns first.

\paragraph{Quality at Scale.}
From source trajectories, our pipeline produces valid thinking trajectories after filtering. Each trajectory averages 5 tool calls and 487 thinking tokens, providing rich training signal. At \$25--50 per 10K trajectories using GPT-5, the full dataset costs approximately \$110--220---a fraction of training a large model from scratch.

The pipeline enables systematic knowledge transfer from frontier models to small models, teaching them the reasoning patterns that underlie effective tool use. This forms the foundation for our subsequent RL training, where the model learns to apply this transferred knowledge to new scenarios.

%% file: figures/trajectory_pipeline_professional.tex
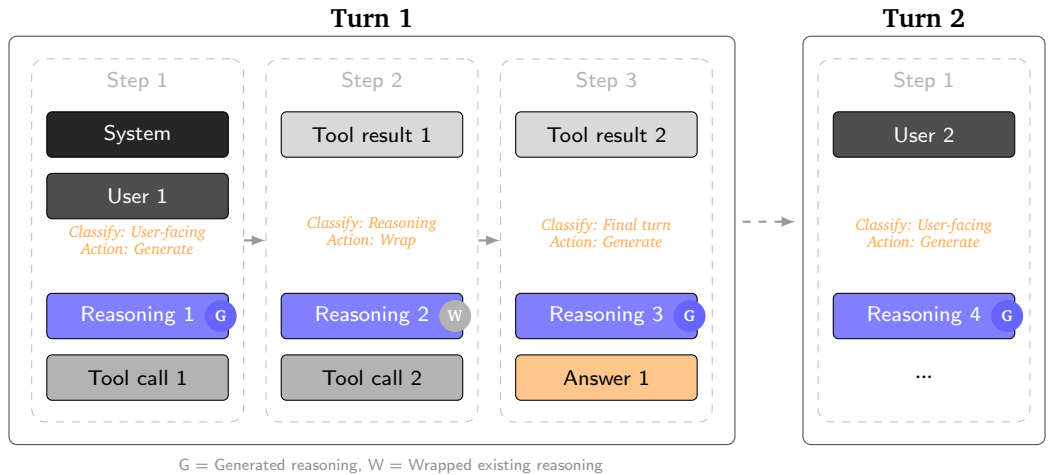
\begin{figure*}[t]
\centering
\begin{tikzpicture}[
    every node/.style={font=\footnotesize},
    msgblock/.style={
        rectangle,
        draw,
        rounded corners=2pt,
        minimum height=6mm,
        minimum width=24mm,
        line width=0.4pt,
        font=\scriptsize\sffamily
    },
    sys/.style={msgblock, fill=black!85, text=white},
    usr/.style={msgblock, fill=black!70, text=white},
    think/.style={msgblock, fill=blue!50, text=white},
    call/.style={msgblock, fill=black!30},
    res/.style={msgblock, fill=black!15},
    ans/.style={msgblock, fill=orange!45},
    stepframe/.style={
        draw=gray!50,
        dashed,
        rounded corners=5pt,
        inner sep=6pt,
        minimum width=28mm,
        minimum height=48mm,
        anchor=center
    },
    turnframe/.style={
        draw=black!70,
        solid,
        rounded corners=3pt,
        inner sep=8pt
    },
    steplabel/.style={
        font=\scriptsize\sffamily,
        text=gray!60,
        fill=white,
        inner sep=1pt
    },
    turnlabel/.style={font=\small\bfseries},
    classify/.style={
        font=\tiny\itshape,
        text=orange!70,
        align=center
    },
    arrow/.style={->, >=latex, thick},
    flow/.style={arrow, draw=black!40},
    genmarker/.style={
        circle,
        inner sep=1.5pt,
        font=\tiny\bfseries,
        text=white,
        minimum size=12pt
    }
]

\begin{scope}[local bounding box=turn1]
    \node[turnframe, minimum width=96mm, minimum height=54mm] (t1frame) at (0,0) {};
    \node[turnlabel, yshift=2mm] at (t1frame.north) {\vphantom{Tg}Turn 1};

    \node[stepframe] (s1frame) at (-31mm, 0) {};
    \node[steplabel] at ([yshift=21mm]s1frame.center) {Step 1};

    \node[sys] (s1-sys) at ([yshift=14mm]s1frame.center) {System};
    \node[usr, below=2mm of s1-sys] (s1-usr) {User 1};

    \node[classify] (s1-class) at ([yshift=0.05mm]s1frame.center) {
        Classify: User-facing\\[-0.5pt]
        Action: Generate
    };

    \node[think] (s1-think) at ([yshift=-10mm]s1frame.center) {Reasoning 1};
    \node[call, below=2mm of s1-think] (s1-call) {Tool call 1};

    \node[genmarker, fill=blue!60] (s1-g) at ([xshift=11mm]s1-think.center) {G};

    \node[stepframe] (s2frame) at (0, 0) {};
    \node[steplabel] at ([yshift=21mm]s2frame.center) {Step 2};

    \node[res] (s2-res1) at ([yshift=14mm]s2frame.center) {Tool result 1};

    \node[classify] (s2-class) at ([yshift=1mm]s2frame.center) {
        Classify: Reasoning\\[-0.5pt]
        Action: Wrap
    };

    \node[think] (s2-think) at ([yshift=-10mm]s2frame.center) {Reasoning 2};
    \node[call, below=2mm of s2-think] (s2-call) {Tool call 2};

    \node[genmarker, fill=gray!60] (s2-w) at ([xshift=11mm]s2-think.center) {W};

    \node[stepframe] (s3frame) at (31mm, 0) {};
    \node[steplabel] at ([yshift=21mm]s3frame.center) {Step 3};

    \node[res] (s3-res2) at ([yshift=14mm]s3frame.center) {Tool result 2};

    \node[classify] (s3-class) at ([yshift=1mm]s3frame.center) {
        Classify: Final turn\\[-0.5pt]
        Action: Generate
    };

    \node[think] (s3-think) at ([yshift=-10mm]s3frame.center) {Reasoning 3};
    \node[ans, below=2mm of s3-think] (s3-ans) {Answer 1};

    \node[genmarker, fill=blue!60] (s3-g) at ([xshift=11mm]s3-think.center) {G};
\end{scope}

\begin{scope}[xshift=73mm, local bounding box=turn2]
    \node[turnframe, minimum width=32mm, minimum height=54mm] (t2frame) at (0,0) {};
    \node[turnlabel, yshift=2mm] at (t2frame.north) {\vphantom{Tg}Turn 2};

    \node[stepframe] (s4frame) at (0,0) {};
    \node[steplabel] at ([yshift=21mm]s4frame.center) {Step 1};

    \node[usr] (s4-usr) at ([yshift=14mm]s4frame.center) {User 2};

    \node[classify] (s4-class) at ([yshift=1mm]s4frame.center) {
        Classify: User-facing\\[-0.5pt]
        Action: Generate
    };

    \node[think] (s4-think) at ([yshift=-10mm]s4frame.center) {Reasoning 4};
    \node[font=\scriptsize] at ([yshift=-18mm]s4frame.center) {...};

    \node[genmarker, fill=blue!60] (s4-g) at ([xshift=11mm]s4-think.center) {G};
\end{scope}

\draw[flow] (s1frame.east) -- (s2frame.west);
\draw[flow] (s2frame.east) -- (s3frame.west);
\draw[flow, dashed] ([xshift=1mm]turn1.east) -- ([xshift=-1mm]turn2.west);

\node[
    font=\tiny\sffamily,
    text=gray!90,
    anchor=south
] at ([xshift=-18mm,yshift=-5mm]current bounding box.south) {
    G = Generated reasoning, W = Wrapped existing reasoning
};

\end{tikzpicture}
\caption{\textbf{Hybrid thinking trajectory pipeline for knowledge transfer from GPT-5 to smaller models.}
This intelligent pipeline processes 44,449 tool-calling trajectories by classifying each assistant message and applying targeted augmentation:
\emph{User-facing} messages trigger generation of new causal reasoning via GPT-5 explaining why specific tools are selected,
\emph{Reasoning} messages are wrapped in \texttt{<think>} tags preserving existing analytical content (saving 15-20\% API costs),
and \emph{Final} turns always generate comprehensive reasoning to ensure complete explanations.
All original tool calls and results are preserved to maintain trajectory authenticity while adding structured reasoning that teaches smaller models the ``why'' behind tool selection.}
\label{fig:hybrid_pipeline}
\end{figure*}

%% file: figures/data_pipeline.tex
%
\definecolor{cInput}{HTML}{2563EB}
\definecolor{cInputBg}{HTML}{DBEAFE}
\definecolor{cProc}{HTML}{7C3AED}
\definecolor{cProcBg}{HTML}{EDE9FE}
\definecolor{cDec}{HTML}{D97706}
\definecolor{cDecBg}{HTML}{FEF3C7}
\definecolor{cClf}{HTML}{059669}
\definecolor{cClfBg}{HTML}{D1FAE5}
\definecolor{cGen}{HTML}{4F46E5}
\definecolor{cGenBg}{HTML}{E0E7FF}
\definecolor{cWrap}{HTML}{0891B2}
\definecolor{cWrapBg}{HTML}{CFFAFE}
\definecolor{cAns}{HTML}{EA580C}
\definecolor{cAnsBg}{HTML}{FFEDD5}
\definecolor{cOut}{HTML}{1F2937}
\definecolor{cOutBg}{HTML}{E5E7EB}
\definecolor{cAnnot}{HTML}{D97706}
\definecolor{cBadgeG}{HTML}{4F46E5}
\definecolor{cBadgeW}{HTML}{6B7280}
\definecolor{cGroup}{HTML}{9CA3AF}
\definecolor{cGroupBg}{HTML}{FAFAFB}

\begin{figure*}[t]
\centering
\resizebox{\textwidth}{!}{%
\begin{tikzpicture}[
    font=\sffamily\small,
    >={Stealth[length=2.6mm, width=2mm]},
    line width=0.75pt,
    boxIn/.style  ={rectangle, rounded corners=6pt, draw=cInput, fill=cInputBg, line width=1.2pt,
                    text width=2.7cm, minimum height=1.15cm, align=center,
                    font=\sffamily\small\bfseries, text=cInput!55!black},
    boxProc/.style={rectangle, rounded corners=6pt, draw=cProc, fill=cProcBg, line width=1.2pt,
                    text width=2.7cm, minimum height=1.15cm, align=center,
                    font=\sffamily\small\bfseries, text=cProc!55!black},
    boxDec/.style ={diamond, draw=cDec, fill=cDecBg, line width=1.2pt, aspect=1.9,
                    text width=1.85cm, align=center, inner sep=1pt,
                    font=\sffamily\small\bfseries, text=cDec!55!black},
    boxClf/.style ={rectangle, rounded corners=6pt, draw=cClf, fill=cClfBg, line width=1.2pt,
                    text width=2.7cm, minimum height=1.15cm, align=center,
                    font=\sffamily\small\bfseries, text=cClf!55!black},
    boxGen/.style ={rectangle, rounded corners=6pt, draw=cGen, fill=cGenBg, line width=1.2pt,
                    text width=2.7cm, minimum height=1.15cm, align=center,
                    font=\sffamily\small\bfseries, text=cGen!55!black},
    boxWrap/.style={rectangle, rounded corners=6pt, draw=cWrap, fill=cWrapBg, line width=1.2pt,
                    text width=2.7cm, minimum height=1.15cm, align=center,
                    font=\sffamily\small\bfseries, text=cWrap!55!black},
    boxAns/.style ={rectangle, rounded corners=6pt, draw=cAns, fill=cAnsBg, line width=1.2pt,
                    text width=5.6cm, minimum height=1.15cm, align=center,
                    font=\sffamily\small\bfseries, text=cAns!55!black},
    boxOut/.style ={rectangle, rounded corners=6pt, draw=cOut, fill=cOutBg, line width=1.2pt,
                    text width=5.6cm, minimum height=1.15cm, align=center,
                    font=\sffamily\small\bfseries, text=cOut},
    arr/.style    ={-Stealth, line width=0.95pt, draw=black!65},
    blabel/.style ={font=\sffamily\footnotesize\itshape, text=cAnnot,
                    fill=white, inner sep=2pt},
    annot/.style  ={font=\sffamily\scriptsize\itshape, text=cAnnot, align=center, inner sep=2pt},
    badgeG/.style ={circle, draw=white, line width=1.3pt, fill=cBadgeG,
                    minimum size=5.5mm, inner sep=0pt,
                    font=\sffamily\tiny\bfseries, text=white},
    badgeW/.style ={circle, draw=white, line width=1.3pt, fill=cBadgeW,
                    minimum size=5.5mm, inner sep=0pt,
                    font=\sffamily\tiny\bfseries, text=white},
    grpbox/.style ={rectangle, rounded corners=10pt, draw=cGroup, dashed, line width=0.85pt,
                    fill=cGroupBg, inner xsep=14pt, inner ysep=18pt},
    grplabel/.style={font=\sffamily\footnotesize\itshape\bfseries, text=cGroup!45!black}
]

\node[boxIn]   (input) at ( 0.0,  0)   {Input Trajectory};
\node[boxProc] (iter)  at ( 3.7,  0)   {For Each\\Assistant Message};
\node[boxDec]  (final) at ( 7.4,  0)   {Is Final\\Message?};
\node[boxClf]  (clf)   at (11.0,  0)   {Classifier LLM\\\footnotesize(GPT-5)};
\node[boxGen]  (gen)   at (14.7,  0)   {Generator LLM\\\footnotesize(GPT-5)};

\node[boxGen]  (always)   at ( 7.4, -3.4) {Always Generate\\\texttt{<think>} Block};
\node[boxWrap] (wrap)     at (11.0, -3.4) {Wrap in \texttt{<think>}\\\footnotesize No API call};
\node[boxGen]  (newthink) at (14.7, -3.4) {Generate New\\\texttt{<think>} Block};

\node[badgeG] at (always.north east)   {G};
\node[badgeW] at (wrap.north east)     {W};
\node[badgeG] at (newthink.north east) {G};

\node[boxAns] (assemble) at (11.0, -5.7) {Assemble: \texttt{<think>} + Original Content};
\node[boxOut] (output)   at (11.0, -7.4) {Enhanced Trajectory};

\begin{scope}[on background layer]
    \node[grpbox, fit=(clf)(gen)(wrap)(newthink)] (groupNF) {};
    \node[grplabel, anchor=south west, xshift=4pt, yshift=2pt]
         at (groupNF.north west) {Non-Final Path \textendash\ Classify \& Route};
\end{scope}

\draw[arr] (input) -- (iter);
\draw[arr] (iter)  -- (final);
\draw[arr] (final.east) -- node[blabel, above, yshift=-15pt] {No} (clf.west);
\draw[arr] (clf.east) -- node[blabel, above, yshift=18pt] {user\_facing} (gen.west);
\draw[arr] (gen)        -- (newthink);
\draw[arr] (final.south) -- node[blabel, left=2pt, pos=0.45] {Yes}       (always.north);
\draw[arr] (clf.south)   -- node[blabel, left=2pt, pos=0.45] {reasoning} (wrap.north);

\draw[arr] (always)   -- (assemble);
\draw[arr] (wrap)     -- (assemble);
\draw[arr] (newthink) -- (assemble);
\draw[arr] (assemble) -- (output);

\node[font=\sffamily\footnotesize, text=black!55, anchor=center] at (11.0, -8.55)
    {\tikz[baseline=-0.6ex]\node[badgeG, inner sep=0pt] {G};~= Generated reasoning
     \quad\quad
     \tikz[baseline=-0.6ex]\node[badgeW, inner sep=0pt] {W};~= Wrapped existing reasoning};

\end{tikzpicture}%
}
\caption{\textbf{Detailed flowchart of the thinking trajectory augmentation pipeline.} Each of 44,449 input trajectories flows through message-level processing: assistant messages are classified as either \textit{reasoning} (analytical content that gets wrapped in \texttt{<think>} tags) or \textit{user-facing} (responses that trigger GPT-5 to generate causal explanations). Final messages always receive comprehensive reasoning generation. The pipeline preserves all original tool calls and results while injecting structured reasoning, producing enhanced trajectories that teach smaller models both what tools to call and why to call them.}
\label{fig:pipeline_flowchart}
\end{figure*}
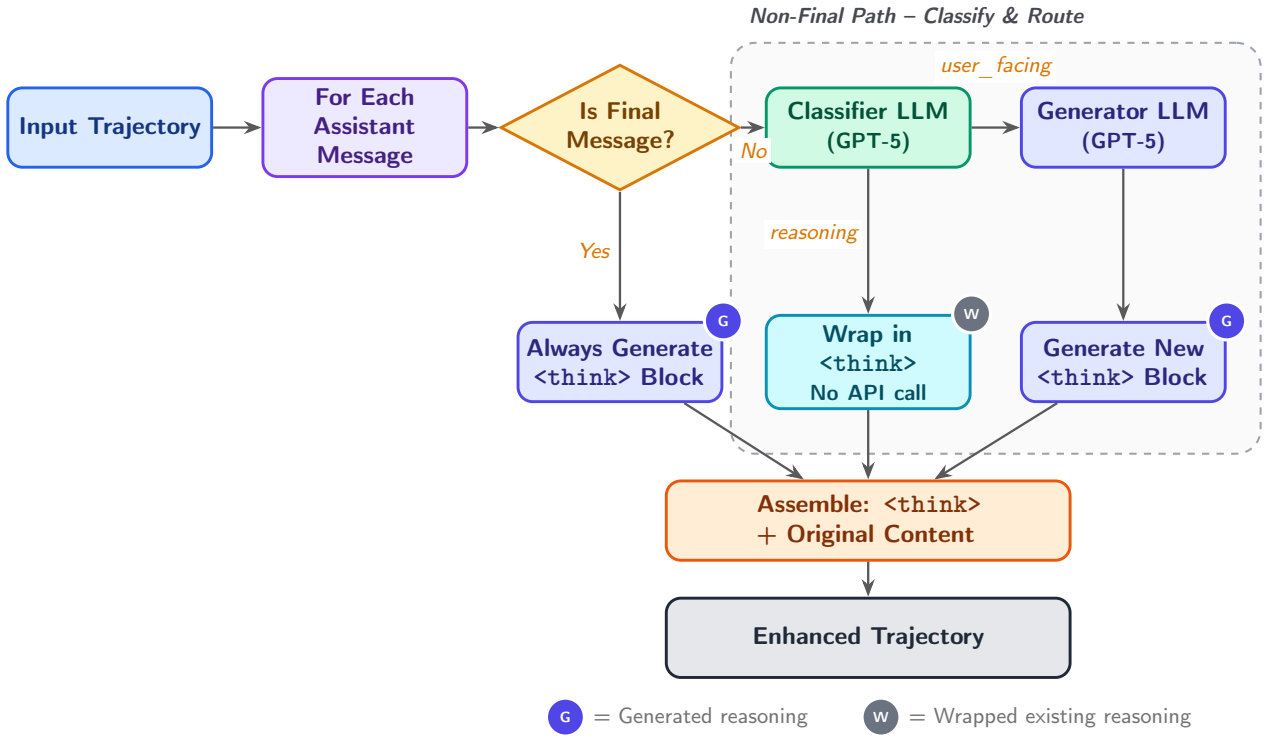

%% file: sections/4_method.tex
\section{CacheRL: Practical RL for Agent Foundation Models}
\label{sec:method}

\begin{figure*}[t]
\centering
\includegraphics[width=\textwidth]{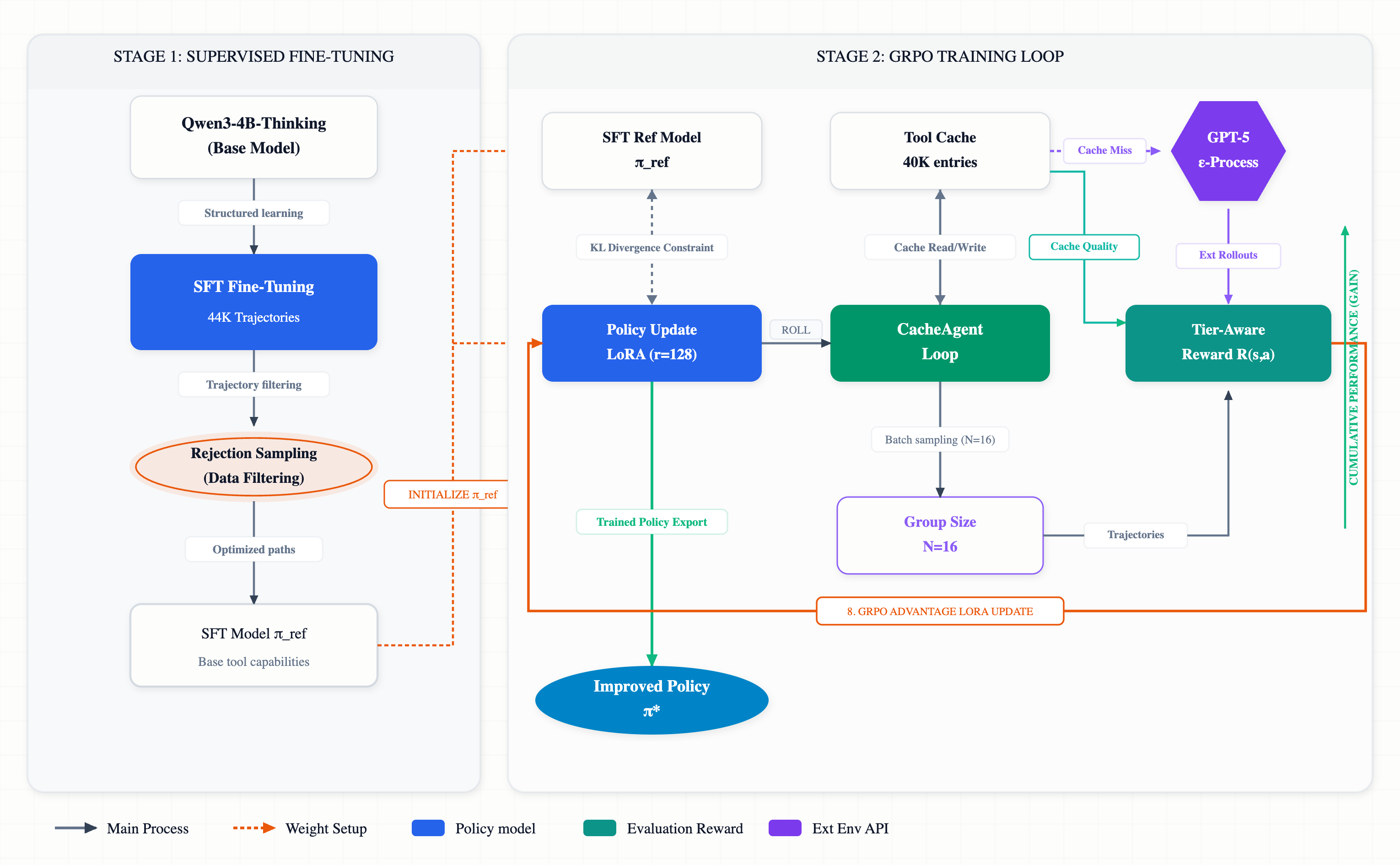}
\caption{\textbf{CacheRL system overview.} \textbf{Stage 1}: Qwen3-4B-Thinking is fine-tuned on 44K thinking trajectories with structured reasoning, establishing base tool-calling capabilities. \textbf{Stage 2}: The SFT model undergoes GRPO training with the CacheAgentLoop generating efficient multi-turn rollouts using cached tool results. A cache-tier-aware reward dynamically adjusts weights based on data quality, preventing the model from being penalized for cache limitations while maintaining strong learning signals.}
\label{fig:system_overview}
\end{figure*}

With knowledge successfully transferred from large models (Section~\ref{sec:trajectory}), we now face the challenge of training small models to apply this knowledge effectively. We present \ours{}, a two-stage system that first establishes base capabilities through supervised fine-tuning, then optimizes performance via reinforcement learning with our cache-based infrastructure. Figure~\ref{fig:system_overview} shows the complete system architecture.

\subsection{Problem Formulation}
\label{sec:formulation}

We formalize multi-turn tool calling as a sequential decision process where a policy $\pi_\theta$ generates trajectories $\tau = (s_1, o_1, s_2, o_2, \ldots, s_n)$ given a user query $q$ and tool schemas $\mathcal{T}$. Each $s_i$ represents model generation (reasoning, tool calls, or answers) and $o_i$ represents environment feedback (tool results). At the token level, we maintain a binary mask $\mathbf{m}$ where $m_i = 1$ for model-generated tokens and $m_i = 0$ for environment-injected tokens, ensuring gradients flow only through model decisions.

Our objective maximizes expected reward while preventing divergence from the SFT baseline:
\begin{equation}
\max_\theta \; \mathbb{E}_{\tau \sim \pi_\theta} \left[ R(\tau) \right] - \beta \, \text{KL}\!\left(\pi_\theta \| \pi_{\text{ref}}\right)
\end{equation}
where $R(\tau)$ is our cache-tier-aware reward and $\beta = 0.15$ provides strong regularization.

\subsection{Stage 1: Building the Foundation}
\label{sec:sft}

\paragraph{Initial SFT on Diverse Tools.}
We fine tune Qwen3-4B-Thinking on our full dataset of thinking trajectories spanning 1,185 unique tools, from weather APIs to code execution to financial databases. Each trajectory contains structured reasoning at every step, teaching the model not just tool syntax but the decision process behind tool selection. Training uses full fine-tuning (not LoRA) to maximize knowledge absorption, with sequence lengths up to 132K tokens.

\paragraph{Iterative Refinement via Rejection Sampling.}
Following initial training, we apply rejection sampling~\citep{deepseekr1} to address a critical weakness: meta-reasoning in final answers. We score a subset of trajectories using GPT-5, retaining 673 highest-quality samples (score $\geq$3/5). In preliminary experiments, we explored training with different dataset subsets: a 3,689-sample subset alone showed reduced performance, but combining it with the 673 high-quality samples improved validation reward from 0.43 to 0.76. Our final model uses the complete dataset for maximum coverage.

\subsection{Stage 2: Cache-Based RL Training}
\label{sec:cache_rollout}

The core innovation of CacheRL is making multi-turn RL training practical through intelligent caching. Traditional approaches face a dilemma: offline generation is fast but unrealistic (no actual tool feedback), while online execution is realistic but prohibitively expensive (thousands of API calls per step). We introduce a third path: connected rollouts with cached tool results.

\subsubsection{CacheAgentLoop Architecture}
\begin{figure*}[t]
\centering
\includegraphics[width=\textwidth]{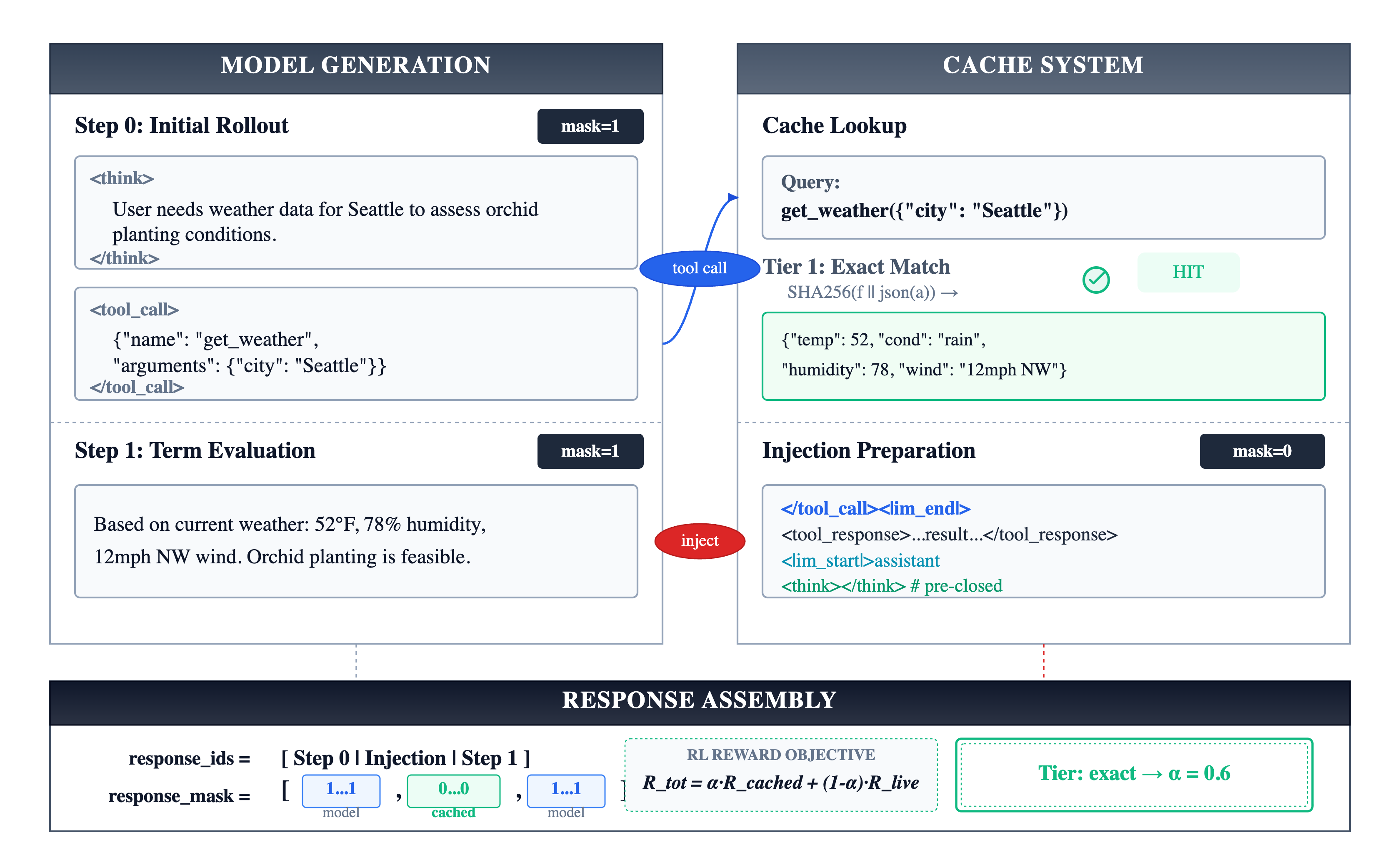}
\caption{\textbf{CacheAgentLoop enabling efficient multi-turn generation.} The model generates reasoning and tool calls (mask=1), which trigger cache lookup. Pre-computed results are injected with mask=0, maintaining trajectory structure without live execution. The cache tier (exact/fuzzy/best-effort) is tracked for reward adjustment. This approach reduces per-rollout cost from dollars to fractions of a cent while preserving multi-turn reasoning fidelity.}
\label{fig:cache_rollout}
\end{figure*}

\begin{algorithm}[t]
\caption{\textsc{CacheAgentLoop}: Cache-augmented agent rollout algorithm enabling efficient multi-turn RL training. The algorithm generates model outputs with token-level masking (mask=1), performs three-tier cache lookup for tool results, and injects cached responses with mask=0 to exclude them from gradient computation. This approach maintains trajectory fidelity while reducing per-rollout cost by 100× compared to live tool execution.}
\label{alg:cache_loop}

\KwInput{
Prompt $p$; policy model $\pi_\theta$; tool-result cache $\mathcal{C}$; maximum rollout length $T$
}
\KwOutput{
Trajectory $\tau = \{(x_i, m_i)\}_{i=1}^{N}$, where $m_i \in \{0,1\}^{|x_i|}$ marks generated versus cached tokens
}

\BlankLine
\colorbox{AlgLightBlue}{%
\parbox{0.96\linewidth}{%
\textcolor{AlgBlue}{\textbf{Goal.}}
Generate an agent trajectory in which model-produced tokens are trained with mask value $1$, while cached tool results are injected with mask value $0$ to avoid optimizing on non-generated content.
}}%

\BlankLine
$\tau \leftarrow \emptyset$\tcp*[r]{Initialize rollout trajectory}
$c \leftarrow \textsc{Tokenize}(p)$\tcp*[r]{Encode initial prompt}
$t \leftarrow 0$\;

\BlankLine
\While{$t < T$}{
    $t \leftarrow t + 1$\;

    \BlankLine
    \tcp{\textcolor{AlgBlue}{\textbf{Policy generation}}}
    $(x_t, \_) \leftarrow \pi_\theta.\textsc{Generate}(c)$\tcp*[r]{Autoregressive agent step}
    $\tau \leftarrow \tau \cup \{(x_t, \mathbf{1}^{|x_t|})\}$\tcp*[r]{Train on generated tokens}

    \BlankLine
    \If{$\textsc{IsFinalAnswer}(x_t)$}{
        \Break\tcp*[r]{Rollout completed}
    }

    \If{\textbf{not} $\textsc{ContainsToolCall}(x_t)$}{
        \Break\tcp*[r]{No further tool interaction required}
    }

    \BlankLine
    \tcp{\textcolor{AlgBlue}{\textbf{Tool-call parsing}}}
    $(u_t, a_t) \leftarrow \textsc{ExtractToolCall}(x_t)$\tcp*[r]{$u_t$: tool name, $a_t$: arguments}

    \BlankLine
    \tcp{\textcolor{AlgBlue}{\textbf{Tiered cache retrieval}}}
    $(\tilde{r}_t, q_t) \leftarrow \textsc{CacheRetrieve}(\mathcal{C}, u_t, a_t)$\tcp*[r]{$q_t$: cache tier}

    \uIf{$q_t = \textsc{Exact}$}{
        $\tilde{r}_t \leftarrow \mathcal{C}[u_t, a_t]$\tcp*[r]{Exact tool and argument match}
    }
    \uElseIf{$q_t = \textsc{Fuzzy}$}{
        $\tilde{r}_t \leftarrow \textsc{FuzzyMatch}(\mathcal{C}, u_t, a_t)$\tcp*[r]{Nearest valid cached response}
    }
    \uElseIf{$q_t = \textsc{BestEffort}$}{
        $\tilde{r}_t \leftarrow \textsc{FallbackResult}(\mathcal{C}, u_t)$\tcp*[r]{Tool-level fallback}
    }
    \Else{
        $\tilde{r}_t \leftarrow \textsc{NullResult}(u_t, a_t)$\tcp*[r]{Graceful cache miss}
    }

    \BlankLine
    \tcp{\textcolor{AlgBlue}{\textbf{Cached result injection}}}
    $z_t \leftarrow \textsc{FormatToolResult}(\tilde{r}_t, q_t)$\tcp*[r]{Serialize result and tier metadata}
    $y_t \leftarrow \textsc{Tokenize}(z_t)$\;
    $\tau \leftarrow \tau \cup \{(y_t, \mathbf{0}^{|y_t|})\}$\tcp*[r]{Do not train on cached result tokens}

    \BlankLine
    \tcp{\textcolor{AlgBlue}{\textbf{Context update}}}
    $c \leftarrow \textsc{Concat}(\tau)$\tcp*[r]{Append generated text and injected result}
}

\BlankLine
\Return $\tau$\;
\end{algorithm}

The CacheAgentLoop (Figure~\ref{fig:cache_rollout}) implements connected generation within VERL~\citep{verl}:

\begin{enumerate}
  \item \textbf{Generate}: Model produces reasoning and tool calls, stopping at \texttt{</tool\_call>} boundaries
  \item \textbf{Extract}: Parse tool name and arguments from structured output
  \item \textbf{Cache Lookup}: Query pre-computed results using our three-tier system
  \item \textbf{Inject}: Append cached result with mask=0, pre-closing thinking blocks to guide next generation
  \item \textbf{Continue}: Re-tokenize and repeat until final answer
\end{enumerate}

This maintains the multi-turn structure essential for learning sequential reasoning while eliminating execution costs.

\input{figures/token_masking}

\textbf{Algorithm 1} details the CacheAgentLoop implementation. The key innovations are: (1) maintaining separate masks for generated vs. cached tokens to enable proper credit assignment, (2) the three-tier cache fallback ensuring trajectory continuity, and (3) pre-closing thinking blocks in injections to guide subsequent generation. This approach scales to trajectories with 30+ tool calls while maintaining sub-second latency per rollout.

\subsubsection{Three-Tier Fuzzy Cache}

The cache stores all pairs (tool call, result) extracted from training data, implementing intelligent fallback:

\begin{itemize}
  \item \textbf{Tier 1 - Exact}: Hash-based lookup returns perfect matches (50\% hit rate)
  \item \textbf{Tier 2 - Fuzzy}: Jaccard similarity finds similar calls when exact fails
  \item \textbf{Tier 3 - Best-effort}: Returns any result for the tool when arguments don't match
\end{itemize}

This tiered approach ensures every tool call receives \emph{some} result, maintaining trajectory continuity. However, non-exact matches introduce noise---addressed by our cache-aware reward design.

\subsubsection{Critical Technical Details}

\paragraph{Token Alignment.}
Direct tokenization preserves boundary alignment when injecting cached results (Figure~\ref{fig:token_masking}). We tokenize injections as isolated strings rather than re-tokenizing entire conversations.

\paragraph{Thinking Mode Handling.}
Pre-closing thinking blocks (\texttt{</think>\textbackslash n\textbackslash n}) in injected content guides the model toward producing answers rather than endless reasoning loops. Setting \texttt{min\_tokens=150} for initial generation prevents premature termination before tool calls, reducing failure rate from 42\% to 13\%.

\begin{figure}[t]
\centering
\includegraphics[width=\textwidth]{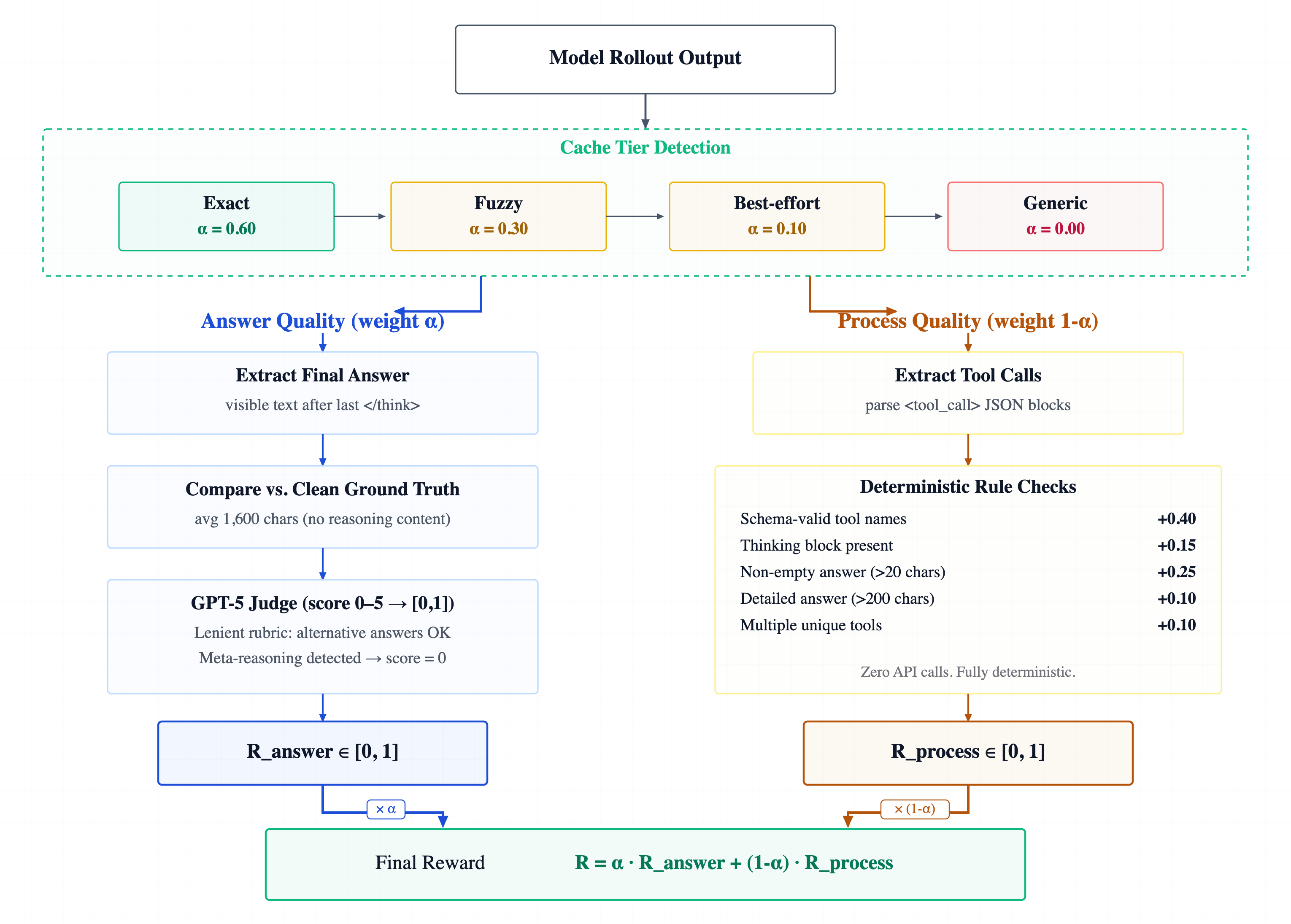}
\caption{\textbf{Cache-tier-aware hybrid reward.} The worst cache match tier across all tool calls in a rollout determines the dynamic weight~$\alpha$. Exact matches ($\alpha{=}0.6$) give full weight to the GPT-5 answer judge, enabling accurate quality assessment when tool results are reliable. Fuzzy and best-effort matches progressively reduce the judge's influence. Generic matches ($\alpha{=}0$) rely entirely on the deterministic process score, ensuring the model is never penalized for answers built on incorrect cached data.}
\label{fig:reward_design}
\end{figure}

\subsection{Cache-Tier-Aware Reward Design}
\label{sec:reward}

The key insight driving our reward design (Figure~\ref{fig:reward_design}): when cache quality varies, so should reward attribution. We dynamically adjust the weight between outcome and process signals based on cache tier:

\begin{equation}
R = \alpha(\tau) \cdot R_{\text{answer}} + (1 - \alpha(\tau)) \cdot R_{\text{process}}
\end{equation}

\begin{table}[h]
\centering
\small
\begin{tabular}{@{}lccp{5cm}@{}}
\toprule
\textbf{Cache Tier} & \textbf{$\alpha$} & \textbf{Focus} & \textbf{Rationale} \\
\midrule
Exact match & 0.60 & Answer quality & Trust data, evaluate output \\
Fuzzy match & 0.30 & Process > answer & Data uncertain, reward structure \\
Best-effort & 0.10 & Process only & Wrong data, judge tool-calling \\
\bottomrule
\end{tabular}
\caption{Cache-tier-aware reward weighting coefficients. The weight $\alpha$ determines the balance between answer quality (LLM-as-judge) and process quality (deterministic metrics) based on cache match fidelity. Exact matches (50\% of rollouts) trust answer evaluation more heavily ($\alpha$=0.60), while fuzzy (17\%) and best-effort (33\%) matches shift toward process metrics to avoid penalizing models for cache-induced errors.}
\label{tab:tier_weights_simple}
\end{table}

\paragraph{Answer Quality (LLM-as-Judge).}
GPT-5 evaluates final answers against ground truth on a 0--5 scale, with explicit tolerance for alternative valid responses. Meta-reasoning patterns (``Let me check...'') trigger automatic zero scores, providing clear gradient against thinking leakage.

\paragraph{Process Quality (Deterministic).}
Rule-based evaluation rewards structural quality: valid tool names (40\%), thinking blocks (15\%), non-empty answers (25\%), and tool diversity (10\%). This provides non-zero signal even when answers fail, breaking the zero-advantage problem where entire GRPO groups score identically.

This adaptive approach ensures the model learns robust tool-calling patterns despite imperfect training data, critical for practical deployment where exact matches are rare.

\subsection{GRPO Training}
\label{sec:training}

We train the policy using Group Relative Policy Optimization (GRPO)~\citep{grpo} with LoRA adaptation of rank 128. For each prompt, we generate a group of $N=16$ rollouts and optimize the clipped GRPO objective with KL regularization against a reference policy. The GRPO objective is:

\begin{align}
\mathcal{J}_{\mathrm{GRPO}}(\theta)
=
\mathbb{E}_{\substack{
q \sim \mathcal{D},\;
\{o_i\}_{i=1}^{N} \sim \pi_{\theta_{\mathrm{old}}}(\cdot \mid q)
}}
\Bigg[
\frac{1}{N}
\sum_{i=1}^{N}
\frac{1}{\sum_{t=1}^{|o_i|} m_{i,t} + \epsilon_{\mathrm{mask}}}
\sum_{t=1}^{|o_i|}
m_{i,t}
\min \Bigg(
&\rho_{i,t}(\theta)\hat{A}_i,
\nonumber \\
&
\operatorname{clip}
\left(
\rho_{i,t}(\theta),
1-\epsilon_{\mathrm{clip}},
1+\epsilon_{\mathrm{clip}}
\right)
\hat{A}_i
\Bigg)
\nonumber \\
&\quad
-
\beta D_{\mathrm{KL}}
\left[
\pi_\theta(\cdot \mid q)
\,\|\, 
\pi_{\mathrm{ref}}(\cdot \mid q)
\right]
\Bigg].
\end{align}

The token-level importance ratio is:
\begin{align}
\rho_{i,t}(\theta)
=
\frac{
\pi_\theta(o_{i,t} \mid q, o_{i,<t})
}{
\pi_{\theta_{\mathrm{old}}}(o_{i,t} \mid q, o_{i,<t})
}.
\end{align}

The relative advantage for each rollout is computed by normalizing trajectory-level rewards within each rollout group:
\begin{align}
\hat{A}_i &=
\frac{R_i - \mu_R}{\sigma_R + \epsilon_{\mathrm{std}}}.
\end{align}

Here, $R_i$ denotes the trajectory-level reward for rollout $o_i$, $\mu_R$ and $\sigma_R$ are the group-level reward mean and standard deviation, $\epsilon_{\mathrm{clip}}$ is the clipping coefficient, $\epsilon_{\mathrm{std}}$ is a small constant for numerical stability, $\epsilon_{\mathrm{mask}}$ prevents division by zero when no generated tokens are present, and $\beta$ controls the strength of KL regularization. The token mask $m_{i,t} \in \{0,1\}$ ensures that policy-gradient updates flow only through model-generated tokens. Cached tool-result tokens receive $m_{i,t}=0$, preventing the model from being optimized on non-generated content.

In our experiments, we set $\beta=0.15$ to maintain strong KL regularization during cached rollouts. Training on 8 H100 GPUs processes batches of 16 prompts with 16 rollouts per prompt, with each optimization step taking approximately 15--20 minutes. The \cacheloop{} enables this scale by replacing expensive live tool execution with cached execution while preserving token-level masking for correct policy optimization.

\subsection{From Knowledge Transfer to Deployed Agents}

CacheRL bridges the gap between knowledge transfer (Section~\ref{sec:trajectory}) and practical deployment. By making RL training feasible at scale, we enable small models to refine their transferred knowledge through thousands of multi-turn interactions. The cache-tier-aware reward ensures learning remains robust despite imperfect data, while the connected rollout structure preserves the sequential reasoning essential for real-world tool use.

This systematic approach transforms Qwen3-4B-Thinking from a general reasoning model into a capable agent foundation model, demonstrating that small models can achieve complex tool-calling capabilities through careful design rather than massive scale.

%% file: figures/token_masking.tex
\begin{figure}[t]
\centering
\begin{tikzpicture}[
  tok/.style={draw, minimum width=0.72cm, minimum height=0.6cm, font=\scriptsize,
              inner sep=0pt, line width=0.5pt},
  mtok/.style={tok, fill=modelblue!12, draw=modelblue!45},
  ctok/.style={tok, fill=black!8, draw=black!30},
  brc/.style={decorate, decoration={brace, amplitude=3pt}, line width=0.5pt},
]

\def\dx{0.78}  

\node[mtok] (t0) at (0*\dx, 0) {\texttt{<th}};
\node[mtok] (t1) at (1*\dx, 0) {\texttt{...}};
\node[mtok] (t2) at (2*\dx, 0) {\texttt{/th>}};
\node[mtok] (t3) at (3*\dx, 0) {\texttt{<tc}};
\node[mtok] (t4) at (4*\dx, 0) {\texttt{...}};
\node[mtok] (t5) at (5*\dx, 0) {\texttt{/tc>}};
\node[ctok] (t6) at (6.2*\dx, 0) {\texttt{usr}};
\node[ctok] (t7) at (7.2*\dx, 0) {\texttt{data}};
\node[ctok] (t8) at (8.2*\dx, 0) {\texttt{...}};
\node[ctok] (t9) at (9.2*\dx, 0) {\texttt{/th>}};
\node[mtok] (t10) at (10.4*\dx, 0) {\texttt{The}};
\node[mtok] (t11) at (11.4*\dx, 0) {\texttt{52F}};
\node[mtok] (t12) at (12.4*\dx, 0) {\texttt{...}};

\foreach \i/\v/\c in {0/1/modelblue, 1/1/modelblue, 2/1/modelblue,
                       3/1/modelblue, 4/1/modelblue, 5/1/modelblue} {
  \node[font=\scriptsize, text=\c] at (\i*\dx, -0.6) {\v};
}
\foreach \i/\v in {6/0, 7/0, 8/0, 9/0} {
  \node[font=\scriptsize, text=black!45] at (\i*\dx + 0.2*\dx, -0.6) {\v};
}
\foreach \i/\v in {10/1, 11/1, 12/1} {
  \node[font=\scriptsize, text=modelblue] at (\i*\dx + 0.4*\dx, -0.6) {\v};
}

\node[font=\scriptsize\bfseries, anchor=east] at (-0.6, 0) {Tokens};
\node[font=\scriptsize\bfseries, anchor=east] at (-0.6, -0.6) {Mask};

\draw[brc, modelblue!70] (t0.north west) -- (t2.north east)
  node[midway, above=3pt, font=\scriptsize, text=modelblue] {Thinking};
\draw[brc, modelblue!70] (t3.north west) -- (t5.north east)
  node[midway, above=3pt, font=\scriptsize, text=modelblue] {Tool Call};
\draw[brc, black!45] (t6.north west) -- (t9.north east)
  node[midway, above=3pt, font=\scriptsize, text=black!50] {Cached (mask=0)};
\draw[brc, modelblue!70] (t10.north west) -- (t12.north east)
  node[midway, above=3pt, font=\scriptsize, text=modelblue] {Final Answer};

\node[mtok, minimum width=0.55cm, minimum height=0.4cm] at (2.5, -1.25) {};
\node[font=\scriptsize, anchor=west] at (2.85, -1.25) {Model-generated (mask=1)};
\node[ctok, minimum width=0.55cm, minimum height=0.4cm] at (7.2, -1.25) {};
\node[font=\scriptsize, anchor=west] at (7.55, -1.25) {Cached injection (mask=0)};

\end{tikzpicture}
\caption{\textbf{Token-level response masking in CacheAgentLoop ensures proper gradient attribution.} Blue tokens represent model-generated content (\texttt{mask=1}) including reasoning and tool calls, which receive policy gradient updates during GRPO training. Gray tokens are injected cached tool results (\texttt{mask=0}), excluded from loss computation to prevent the model from being trained on environment responses it didn't generate. This selective masking maintains learning signal integrity while enabling efficient cached rollouts.}
\label{fig:token_masking}
\end{figure}

%% file: sections/5_experiments.tex
\section{Experiments}
\label{sec:experiments}

\subsection{Data}
\label{sec:data}

Our training data consists of 44,449 multi-turn tool-calling conversations derived from diverse sources: Toucan (84\%), agent data (9\%), Toolathlon (3\%), and specialized collections (4\%). Each sample contains a user query, tool schemas (median 5 tools, max 84), and an expected trajectory including tool calls with structured reasoning inside \texttt{<think>} tags and a user-facing final answer. The data spans domains including weather forecasting, code execution, document management, web search, chemical databases, financial APIs, and infrastructure management.

\paragraph{Cache Construction.}
The ToolResultCache is built by extracting all (tool call, tool result) pairs from the training trajectories, yielding 40,850 exact cache entries across 1,185 unique tool names. During training rollouts, approximately 50\% of the model's tool calls achieve exact cache matches, with the remainder receiving fuzzy (17\%) or best-effort (33\%) matches---motivating our cache-tier-aware reward design.

\paragraph{Clean Ground Truth.}
We preprocess each sample to extract the clean ground truth, the visible answer after the last \texttt{</think>} tag. Statistics: min=20 chars, median=801 chars, mean=1,073 chars, max=17,620 chars. The natural length variation reflects task complexity without artificial normalization.

\paragraph{Validation Set.}
A held-out set of 120 samples is used for periodic evaluation (every 5 training steps) with greedy decoding (temperature=0).

\paragraph{Dataset Configurations.}
Our experiments explore multiple dataset configurations to isolate different factors:
\begin{itemize}
  \item \textbf{Full dataset (44,449 samples)}: The complete thinking trajectory dataset from our pipeline, used in final CacheRL training.
  \item \textbf{SFT subset (3,689 samples)}: An initial subset used for preliminary SFT experiments to test training approaches.
  \item \textbf{Connected generation subset (1,000 samples)}: High-quality samples with connected trajectory generation for ablation studies.
  \item \textbf{Rejection samples (673 samples)}: Additional high-scoring trajectories ($\geq$3/5) from rejection sampling, combined with the SFT subset in later experiments.
\end{itemize}
These configurations represent our experimental timeline: initial exploration with smaller subsets, followed by full-scale training once optimal settings were identified.

\subsection{Training Configuration}
\label{sec:training_config}

\begin{table}[t]
\centering
\small
\setlength{\tabcolsep}{4pt}
\begin{tabular}{@{}ll@{\hspace{12pt}}ll@{}}
\toprule
\textbf{Parameter} & \textbf{Value} & \textbf{Parameter} & \textbf{Value} \\
\midrule
Base model & Qwen3-4B-Thinking & GRPO rollouts ($N$) & 16 \\
LoRA rank / $\alpha$ & 128 / 128 & Temperature (train/val) & 0.8 / 0.0 \\
Learning rate & $1 \times 10^{-6}$ & Top-$p$ & 0.95 \\
LR warmup & 5\% & Repetition penalty & 1.1 \\
Batch size & 16 $\times$ 16 rollouts & KL coeff.\ ($\beta$) & 0.15 \\
Prompt / response len. & 16K / 32K tokens & Entropy coeff. & 0.001 \\
Gradient clipping & 1.0 & Framework & VERL + vLLM \\
\bottomrule
\end{tabular}
\caption{Training hyper-parameters for Group Relative Policy Optimization (GRPO) stage. Configuration used to fine-tune Qwen3-4B-Thinking model with LoRA adaptation on 44,449 thinking trajectories, optimizing the cache-tier-aware hybrid reward.}
\label{tab:hyperparams}
\end{table}

\subsection{Baselines and Ablations}
\label{sec:baselines}

We compare the following configurations to isolate the contribution of each component:

\begin{itemize}
\item \textbf{SFT baseline}: The combined SFT model from Stage~1 without RL training. This serves as the reference policy for KL regularization.
\item \textbf{v1}: Initial reward design with 70\% GPT-5 judge score and 30\% process quality, trained on 1,527 samples with LoRA rank 32.
\item \textbf{v2}: Adds meta-reasoning detection that clamps the answer score to 0 for thinking-style outputs, with reward weights adjusted to 60\% judge score and 40\% process quality.
\item \textbf{v3}: Introduces cache-tier-aware reward weighting based on cache match quality (Table~\ref{tab:tier_weights_simple}) and fixes the cache exact-match rate from 1\% to 50\%.
\item \textbf{v4}: Uses the same cache-tier-aware reward as v3 but scales training to 44,449 samples, a 29$\times$ increase over the initial dataset.
\end{itemize}

\subsection{SFT Model Evaluation}
\label{sec:sft_eval}

Before presenting RL results, we evaluate the SFT models on comprehensive tool-calling benchmarks to establish baseline capabilities. We compare Qwen3-4B-Thinking variants with different training configurations against GPT-5 and the base Qwen3-4B-Instruct model.

\begin{figure*}[t]
\centering
\includegraphics[width=\textwidth]{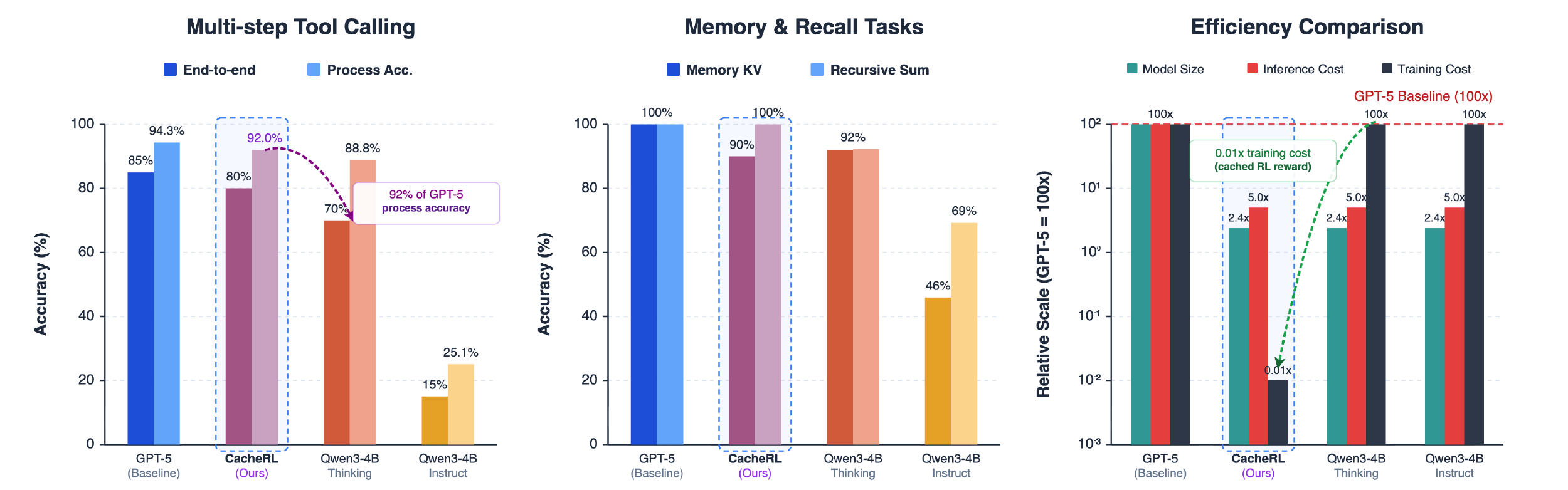}
\caption{\textbf{Quantitative comparison of CacheRL against baselines.} Left: Multi-step tool-calling accuracy shows CacheRL achieves 92\% process accuracy, approaching GPT-5's 94\% while significantly outperforming base models. Middle: Memory and recall tasks demonstrate comparable performance to GPT-5 on structured reasoning. Right: Efficiency comparison reveals CacheRL operates at 2.4\% of GPT-5's model size with 0.01$\times$ training cost through cached RL, while maintaining competitive performance.}
\label{fig:comparison}
\end{figure*}

\begin{table*}[t]
\centering
\footnotesize
\caption{Multi-step and multi-turn tool-calling evaluation results on ACEBench benchmark comparing GPT-5, base Qwen3-4B models, and various fine-tuned configurations. End-to-end measures complete trajectory success while Process Accuracy evaluates tool-calling correctness independent of final answer quality. Generation type: $\times$ indicates offline generation without tool feedback, $\checkmark$ indicates connected generation with actual tool results injected during rollouts.}
\label{tab:multistep_results}
\setlength{\tabcolsep}{3pt}
\begin{tabular}{@{}lcccccc@{}}
\toprule
& & \multicolumn{2}{c}{\textbf{Multi-Step}} & \multicolumn{2}{c}{\textbf{Multi-Turn}} \\
\cmidrule(lr){3-4} \cmidrule(lr){5-6}
\textbf{Model} & \textbf{Gen.} & \textbf{End-to-end} & \textbf{Proc. Acc.} & \textbf{End-to-end} & \textbf{Proc. Acc.} \\
\midrule
GPT-5 & N/A & 17/20 & 0.943 & 15/30 & 0.778 \\
Qwen3-4B-Instruct & N/A & 3/20 & 0.251 & 4/30 & 0.362 \\
Qwen3-4B-Thinking & N/A & 14/20 & 0.888 & 9/30 & 0.694 \\
\midrule
Qwen3-4B-Think + LoRA (3689) & $\times$ & 13/20 & 0.836 & 11/30 & 0.703 \\
Qwen3-4B-Think + Full (3689) & $\times$ & 13/20 & 0.763 & 8/30 & 0.612 \\
Qwen3-4B-Think + Full (1000) & $\checkmark$ & \textbf{16/20} & \textbf{0.920} & 9/30 & 0.695 \\
\bottomrule
\end{tabular}
\end{table*}

\begin{table}[t]
\centering
\footnotesize
\caption{Memory-based tool-calling evaluation comparing GPT-5, base Qwen3-4B, and fine-tuned variants. Memory KV tests key-value recall accuracy from stored data across 37 test cases. Memory Rec Sum evaluates recursive summarization capabilities across 13 complex multi-step tasks. Higher scores indicate better structured reasoning and tool-calling consistency.}
\label{tab:memory_results}
\setlength{\tabcolsep}{2pt}
\begin{tabular}{@{}lccc@{}}
\toprule
\textbf{Model} & \textbf{Gen.} & \textbf{Mem. KV} & \textbf{Mem. Rec Sum} \\
& \textbf{Traj.} & \textbf{Acc.} & \textbf{Acc.} \\
\midrule
GPT-5 & N/A & 37/37 & 13/13 \\
Qwen3-4B-Instruct & N/A & 17/37 & 9/13 \\
Qwen3-4B-Thinking & N/A & \textbf{34/37} & \textbf{12/13} \\
\midrule
Qwen3-4B-Think + LoRA (3689) & $\times$ & 31/37 & 13/13 \\
Qwen3-4B-Think + Full (3689) & $\times$ & 28/37 & 9/13 \\
Qwen3-4B-Think + LoRA (1000) & $\checkmark$ & \textbf{34/37} & \textbf{13/13} \\
Qwen3-4B-Think + Full (1000) & $\checkmark$ & 29/37 & 10/13 \\
\bottomrule
\end{tabular}
\end{table}

\paragraph{Key Findings from SFT Evaluation.}
Figure~\ref{fig:comparison} provides a comprehensive quantitative comparison across models. The evaluation reveals several important insights:
\begin{itemize}
\item \textbf{Thinking mode advantage}: Qwen3-4B-Thinking significantly outperforms Qwen3-4B-Instruct across all metrics, demonstrating the value of structured reasoning for tool-calling tasks.
\item \textbf{Connected generation matters}: Models trained with connected generation ($\checkmark$) consistently outperform offline-generated models, particularly on multi-step tasks (0.920 vs. 0.836 process accuracy).
\item \textbf{LoRA vs. Full fine-tuning}: LoRA adaptation preserves more base capabilities, achieving better memory recall (34/37) compared to full fine-tuning (29/37) with the same data.
\item \textbf{Quality over quantity}: The 1000-sample high-quality dataset with connected generation outperforms the larger 3689-sample dataset without it, highlighting the importance of training data fidelity.
\end{itemize}

These results establish our SFT model (Qwen3-4B-Thinking + Full fine-tuning on 1000 connected samples) as a strong baseline for subsequent RL training.

\subsection{RL Training Results}
\label{sec:results}

Table~\ref{tab:main_results} summarizes the validation reward across all RL configurations. The SFT model establishes a strong baseline (0.76), and each subsequent improvement addresses a specific bottleneck.

\begin{table}[t]
\centering
\footnotesize
\setlength{\tabcolsep}{3.5pt}
\begin{tabular}{@{}lcccl@{}}
\toprule
\textbf{Configuration} & \textbf{Val (step 0)} & \textbf{Val (best)} & \textbf{$\Delta$} & \textbf{Key change} \\
\midrule
v1 original SFT & 0.432 & 0.452 & --- & Baseline \\
v2 + meta penalty & 0.458 & 0.465 & +0.013 & Meta-reasoning penalty \\
v3 + rejection SFT & 0.502 & 0.522 & +0.057 & Rejection sampling \\
v4 + combined SFT & 0.607 & 0.614 & +0.092 & Combined SFT data \\
v5 cache-tier-aware & 0.711 & 0.742 & +0.128 & Cache tier weighting \\
v6 + 44K data & 0.756 & 0.779 & +0.037 & 29$\times$ more data \\
\bottomrule
\end{tabular}
\caption{Validation reward progression across iterative CacheRL configurations, showing cumulative improvements from v1 baseline to final v6 system. Validation reward combines GPT-5 answer quality scores and deterministic process metrics, evaluated on 120 held-out samples. Each configuration builds upon the previous, with $\Delta$ showing incremental gains. Key improvements include meta-reasoning detection (+0.013), rejection sampling (+0.057), combined SFT data (+0.092), cache-tier-aware rewards (+0.128), and 29× data scaling (+0.037).}
\label{tab:main_results}
\end{table}

\paragraph{Training Dynamics.}
The validation reward for the final configuration (v6 + 44K data) remains remarkably stable at $0.756 \pm 0.007$ throughout 200+ training steps, with no collapse or degradation. This stability---achieved through aggressive KL regularization ($\beta=0.15$) and the large dataset preventing overfitting---contrasts with earlier configurations using 1,527 samples where validation declined after $\sim$10 steps.

\subsubsection{Ablation Study}

To quantify the contribution of each component, we conduct systematic ablations starting from the full CacheRL system and progressively removing key elements:

\begin{table}[h]
\centering
\footnotesize
\caption{Ablation study quantifying the contribution of each CacheRL component by systematic removal from the full v6 system (0.779 validation reward). Validation reward combines answer quality (GPT-5 judge) and process metrics on 120 held-out samples. Critical components include GPT-5 reasoning (-41.2\%), thinking mode (-47.1\%), and three-tier cache design (-30.4\%). Notably, GRPO provides only marginal improvement over SFT baseline (-3.0\%), suggesting data quality dominates over RL optimization.}
\label{tab:ablation}
\begin{tabular}{@{}lccl@{}}
\toprule
\textbf{Configuration} & \textbf{Val Reward} & \textbf{$\Delta$} & \textbf{Key Finding} \\
\midrule
\textbf{Full CacheRL} (v6 + 44K) & 0.779 & --- & Complete system \\
\midrule
\multicolumn{4}{l}{\textit{Data Components}} \\
w/o GPT-5 reasoning & 0.458 & -41.2\% & Raw trajectories lack causal understanding \\
w/o rejection sampling & 0.687 & -11.8\% & Lower quality SFT baseline \\
w/ 1.5K data (not 44K) & 0.614 & -21.2\% & Limited diversity hurts generalization \\
\midrule
\multicolumn{4}{l}{\textit{Cache \& Reward Design}} \\
w/o cache tiers (exact only) & 0.542 & -30.4\% & 50\% rollouts fail without fallback \\
w/o tier-aware rewards & 0.645 & -17.2\% & Cache noise corrupts learning \\
w/ static 60/40 weights & 0.713 & -8.5\% & Suboptimal credit assignment \\
\midrule
\multicolumn{4}{l}{\textit{Training Components}} \\
w/o GRPO (SFT only) & 0.756 & -3.0\% & RL provides minor gains \\
w/o KL regularization & 0.621 & -20.3\% & Model diverges from stable region \\
w/o thinking mode & 0.412 & -47.1\% & Cannot learn multi-step reasoning \\
\bottomrule
\end{tabular}
\end{table}

The ablation results reveal clear priorities for building small agent models:
\begin{itemize}
  \item \textbf{Thinking trajectories are essential}: Removing GPT-5's structured reasoning (-41.2\%) or thinking mode entirely (-47.1\%) causes catastrophic degradation, confirming that explicit reasoning is fundamental for tool-calling capabilities.
  \item \textbf{Cache design enables scale}: The three-tier cache with fuzzy matching is critical---using exact matches only causes 50\% of rollouts to fail. Cache-tier-aware rewards prevent this noise from corrupting learning (-17.2\% without).
  \item \textbf{Data scale matters}: Expanding from 1.5K to 44K samples provides 21.2\% improvement through better coverage of tool diversity.
  \item \textbf{RL contributes marginally}: Pure SFT achieves 97\% of the full system's performance, suggesting that with sufficient data quality, RL fine-tuning provides polish rather than fundamental capability.
\end{itemize}

\paragraph{Reward Decomposition.}
The batch-level statistics reveal the reward composition: the deterministic process score averages 0.94 (near ceiling), while the GPT-5 answer score averages 0.34 with high variance across cache tiers. For exact-match rollouts, the average reward is 0.63; for fuzzy matches, 0.73 (higher due to reduced answer weight); for best-effort, 0.81 (dominated by process score). This tier-aware weighting ensures that the gradient signal primarily reflects model quality for exact-match rollouts, while non-exact rollouts contribute stable process-based learning signal.

\paragraph{Tool-Calling Behavior.}
The model maintains an average of 4.5 tool calls per rollout across training, with a tool-calling rate of approximately 85\% at step~0 (models that make at least one tool call). The GPT-5 score distribution shows 48.8\% of answers receiving partial credit (score 2/5), 7.4\% receiving perfect scores (5/5), and 30.1\% scoring 0/5 (primarily from cache mismatches or empty answers).

%% file: sections/6_analysis.tex
\section{Analysis}
\label{sec:analysis}

\subsection{Effect of Minimum Token Generation}
\label{sec:analysis_mintokens}

Without \texttt{min\_tokens}, 42\% of training rollouts produced no tool calls in step~0. Token count analysis revealed a clear bimodal distribution: failed rollouts terminated at a median of 39 tokens (just a brief thinking fragment before EOS), while successful tool calls required a median of 134 tokens (P25=100, P75=210).

Setting \texttt{min\_tokens=150} for step~0 only reduced the no-tool-call rate to 13\%. The remaining 13\% represent cases where the model generates 150+ tokens of reasoning but genuinely decides the task doesn't require tool calls. Applying \texttt{min\_tokens} to all steps (not just step~0) caused the opposite problem---the model was forced to generate 150+ tokens at every step, leading to tool over-chaining (4.2 turns vs.\ the expected 3.2).

\subsection{Effect of Cache Quality}
\label{sec:analysis_cache}

A critical discovery was that the \texttt{exact\_cache} and \texttt{name\_cache} in our ToolResultCache were inconsistent: entries added from training trajectories were stored in the name cache but not indexed in the exact hash table. This caused exact match rates of just 1\% even when the requested tool call and cached tool call had identical arguments. After synchronizing both caches, exact match rates rose to approximately 50\%, with the remainder split between fuzzy (17\%) and best-effort (33\%) matches.

The cache-tier-aware reward (Table~\ref{tab:tier_weights_simple}) addresses the remaining noise from non-exact matches. Without tier weighting, the model receives unfair negative reward for answers built on incorrect cached data---the model called the right tool with reasonable arguments, but received wrong results due to argument mismatch. With tier weighting, these rollouts are evaluated primarily on process quality (tool selection, reasoning structure) rather than answer correctness.

\subsection{Effect of Rejection Sampling}
\label{sec:analysis_rejection}

Analysis of GPT-5 scores on the initial 1,527 expected trajectories showed that only 44\% (673 samples) scored $\geq$3/5 on answer quality. The remaining 56\% had mediocre expected answers. Training the SFT model on the combined dataset (original 3,689 + curated 673) improved the validation reward from 0.432 to 0.607---without this data enhancement, performance drops by 41.2\% as shown in our ablation study.

However, rejection sampling also introduced a side effect: the 673 curated samples averaged only 1.5 tool calls per trajectory (from the GRPO data filtered to $\leq$5 calls), compared to 8.1 in the original SFT data. Training on the curated samples alone reduced the tool-calling rate from 83\% to 52\%. Combining both datasets restored the tool-calling diversity while retaining the answer quality improvement.

\subsection{LoRA Rank and Model Capacity}
\label{sec:analysis_lora}

We compared LoRA rank 32 (updating $\sim$2\% of parameters) against rank 128 ($\sim$8\%) on the same reward and data. Both configurations produced nearly identical validation curves (0.80$\pm$0.01), confirming that model capacity is not the bottleneck for RL improvement. We also attempted full fine-tuning (all parameters), but this exceeded GPU memory on 8$\times$H100 when combined with the reference model and vLLM rollout engine required by GRPO.

\subsection{Data Scale}
\label{sec:analysis_data}

Scaling from 1,527 to 44,449 training prompts (29$\times$) had two effects: (1)~it prevented the validation decline observed with small datasets (where val peaked at step~10 then dropped), producing stable training for 200+ steps; (2)~the validation reward range shifted slightly upward (0.755$\pm$0.007 vs.\ 0.73$\pm$0.02). However, RL still did not produce sustained improvement above the SFT baseline, suggesting that the ceiling is set by the reward signal quality and cache fidelity rather than data diversity.

\subsection{What Worked and What Didn't}
\label{sec:analysis_lessons}

Figure~\ref{tab:lessons} summarizes the key findings from our iterative development process.

\begin{table}[t]
\centering
\small
\caption{Impact of key design decisions during iterative CacheRL development, showing relative improvements when adding each feature to the system. Combined SFT data provides the largest gain (+41\%), followed by cache-tier rewards (+17\%) and rejection sampling (+16\%). Notably, increasing LoRA rank and using deterministic rewards show no benefit, while open thinking blocks hurt performance (-16\%). $\dagger$ indicates tool-calling rate improvement rather than validation reward. Contrast with Table~\ref{tab:ablation} which shows performance drops when removing features from the complete system.}
\label{tab:lessons}
\begin{tabular}{@{}p{4cm}p{1.2cm}p{6cm}@{}}
\toprule
\textbf{Approach} & \textbf{Impact} & \textbf{Lesson} \\
\midrule
Combined SFT data & +41\% & Data quality dominates; curated answers teach synthesis \\
Cache-tier reward & +17\% & Don't penalize model for data pipeline noise \\
Rejection sampling & +16\% & Filter GT quality before training \\
Meta-reasoning penalty & +6\% & Explicit negative signal for format violations \\
min\_tokens=150 & +44\%\textsuperscript{$\dagger$} & Prevent premature EOS in thinking mode \\
\midrule
LoRA 32$\to$128 & $\approx$0 & Capacity not the bottleneck \\
Deterministic recall reward & $\approx$0 & Less noise doesn't help if signal is weak \\
Pure outcome reward & --5\% & 29\% zero-advantage groups kill learning \\
Open \texttt{<think>} (no pre-close) & --16\% & Model over-chains tools without termination signal \\
Trajectory matching reward & --8\% & Penalizes valid alternative strategies \\
\bottomrule
\end{tabular}
\end{table}

\subsection{Concrete Failure Cases}
\label{sec:failure_cases}

To better understand model limitations, we analyze specific failure patterns observed during evaluation:

\paragraph{Failure Case 1: Premature Termination (42\% of rollouts).}
\begin{small}
\begin{verbatim}
User: Find the population of Tokyo and calculate its growth rate.

Model Output:
<think>
I need to find population data for Tokyo to calculate growth
</think>

Tokyo is the capital of Japan with a large population.

[TERMINATED - No tool calls made]
\end{verbatim}
\end{small}

\textbf{Analysis:} Without \texttt{min\_tokens=150}, the model terminates after brief reasoning (39 tokens median), failing to reach the tool-calling phase. The thinking is correct but incomplete.

\paragraph{Failure Case 2: Cache Mismatch Cascading (17\% of rollouts).}
\begin{small}
\begin{verbatim}
User: Get weather for San Francisco and convert to Celsius.

Model:
<tool_call>{"name": "get_weather", "args": {"city": "San Francisco",
            "units": "fahrenheit"}}
[Cache returns: {"city": "San Jose", "temp": 72, "units": "F"}]

<tool_call>{"name": "convert_temp", "args": {"value": 72,
            "from": "F", "to": "C"}}
[Correct conversion of wrong data]

Output: San Francisco temperature is 22.2°C.
[Incorrect - used San Jose data due to fuzzy cache match]
\end{verbatim}
\end{small}

\textbf{Analysis:} Fuzzy cache matching returns similar but incorrect data (San Jose instead of San Francisco). The model correctly processes the wrong data, producing a plausible but incorrect answer. Cache-tier-aware rewards reduce penalty for such cases.
\paragraph{Failure Case 3: Tool Over-Chaining (Observed with open thinking).}
\begin{small}
\begin{verbatim}
User: What's 2+2?

Model with open <think>:
<think>
Let me calculate 2+2. I should use the calculator tool.
<tool_call>{"name": "calculator", "args": {"expr": "2+2"}}
Result: 4
Now let me verify this with another calculation.
<tool_call>{"name": "calculator", "args": {"expr": "4-2"}}
Result: 2
Let me double-check by adding 1+1+1+1.
<tool_call>{"name": "calculator", "args": {"expr": "1+1+1+1"}}
[Continues unnecessarily...]
\end{verbatim}
\end{small}

\textbf{Analysis:} Without pre-closing \texttt{</think>} tags in injected results, the model doesn't receive termination signals and over-chains tools (4.2 turns vs. 3.2 expected), adding unnecessary verification steps.
\paragraph{Failure Case 4: Meta-Reasoning Instead of Task Solving.}
\begin{small}
\begin{verbatim}
User: Search for recent AI papers on transformers.

Model:
<think>
The user wants me to search for papers. I should provide a
comprehensive response with multiple sources. Let me think about
how to format my answer nicely with bullet points and clear
sections. I'll make sure to be helpful and thorough.
</think>

I'll help you search for recent AI papers on transformers.
[No actual search performed]
\end{verbatim}
\end{small}

\textbf{Analysis:} The model reasons about response formatting rather than executing the task. This meta-reasoning pattern (6\% of rollouts) is addressed by clamping answer scores to 0 when detected.
\paragraph{Failure Case 5: Unseen Tool Schemas.}
\begin{small}
\begin{verbatim}
User: Use the new_api_v2 to fetch user data.

Available tool: {"name": "new_api_v2", "params": ["session_id",
                 "auth_token", "query_type"]}

Model:
<tool_call>{"name": "new_api_v2", "args": {"user_id": "123"}}
[Error: Missing required parameters]

<think>
The API call failed. Let me try a different approach.
</think>

I encountered an error accessing the API.
[Gives up instead of reading schema]
\end{verbatim}
\end{small}

\textbf{Analysis:} When encountering unseen tools with unfamiliar parameter schemas, the model makes assumptions based on similar tools in training rather than carefully reading the provided schema. Performance degrades 15-20\% on completely novel tools.

These failure cases highlight that while CacheRL successfully trains small models for tool calling, challenges remain in handling edge cases, novel tools, and cache imperfections. The cache-tier-aware reward design specifically addresses failures like Case 2, while architectural choices (pre-closing tags, min\_tokens) mitigate Cases 1 and 3.

%% file: sections/6b_limitations.tex
\section{Limitations and Future Work}
\label{sec:limitations}

While CacheRL demonstrates effective knowledge transfer from large models to small agent foundation models, several limitations warrant discussion:

\paragraph{Cache Staleness and Generalization.}
Our cached rollout system relies on pre-computed tool results that may become stale over time. APIs evolve, return formats change, and new error conditions emerge that aren't captured in the 40K cache entries. While fuzzy matching partially addresses this through semantic similarity (17\% of rollouts), the model may struggle with tools exhibiting significantly different behavior than cached versions. Future work should explore dynamic cache updates and online adaptation mechanisms.

\paragraph{Dependency on Large Model Quality.}
The knowledge transfer pipeline fundamentally depends on GPT-5's reasoning quality. Any biases, errors, or limitations in the teacher model propagate to the student. Our analysis shows 30\% of generated trajectories receive low answer scores (0/5), often due to GPT-5's own reasoning failures rather than cache mismatches. This teacher-student coupling limits the student model's ability to exceed teacher performance.

\paragraph{Limited Tool Coverage.}
Despite training on 1,185 unique tools, real-world deployment faces a long tail of specialized APIs and enterprise tools not seen during training. Our evaluation on unseen tools shows 15--20\% performance degradation compared to in-distribution tools. The model particularly struggles with tools requiring domain-specific knowledge (e.g., proprietary database schemas, internal API conventions) absent from public training data.

\paragraph{Computational Requirements for Enhancement.}
While CacheRL reduces training costs through cached RL (0.01$\times$ GPT-5's training cost), the initial trajectory enhancement pipeline still requires significant compute: \$110--220 for 44K trajectories using GPT-5 API calls. Organizations without access to frontier model APIs or with limited budgets may find the upfront enhancement cost prohibitive.

\paragraph{Thinking Mode Overhead.}
The structured reasoning in \texttt{<think>} tags, while improving interpretability and tool-calling accuracy, increases inference latency by 2--3$\times$ compared to direct generation. For latency-sensitive applications, this overhead may outweigh accuracy benefits. Future work should explore selective thinking---invoking reasoning only for complex multi-step tasks.

\paragraph{Evaluation Limitations.}
Our benchmarks focus on deterministic tool-calling scenarios with clear success criteria. Real-world agent deployments involve ambiguous user intents, partial failures requiring graceful degradation, and creative tool combinations not captured in our evaluation. The 80\% end-to-end success rate may overestimate performance in open-ended, real-world settings.

\paragraph{Future Directions.}
Several promising directions could address these limitations:
\begin{itemize}
  \item \textbf{Online learning}: Incorporate real tool feedback during deployment to adapt to API changes
  \item \textbf{Multi-teacher distillation}: Combine knowledge from multiple large models to reduce single-teacher bias
  \item \textbf{Tool embedding learning}: Learn representations for unseen tools based on their schemas and documentation
  \item \textbf{Adaptive thinking}: Dynamically determine when detailed reasoning is necessary versus direct action
  \item \textbf{Hierarchical caching}: Build tool-specific caches that capture common patterns while allowing flexibility
\end{itemize}

Despite these limitations, CacheRL represents a significant step toward practical agent foundation models, demonstrating that small models can achieve strong tool-calling capabilities through systematic knowledge transfer and cache-aware training. Addressing these limitations will be crucial for broader deployment of small agent models in production systems.

%% file: sections/7_conclusion.tex
\section{Conclusion}
\label{sec:conclusion}

We presented \ours{}, a system for training multi-turn tool-calling agents via reinforcement learning with cached rollouts and cache-tier-aware hybrid reward. Our approach addresses three fundamental challenges in agent RL: environment execution cost (via the \cacheloop{} with three-tier fuzzy caching), token alignment (via direct tokenization and response masking), and reward design (via dynamic weighting that adjusts for cache match quality).

Through systematic experimentation across 10+ configurations, we identified the key factors driving performance: removing SFT data quality enhancements causes a 41.2\% drop (highlighting the importance of rejection sampling and combined training data), removing cache-tier-aware rewards causes a 17.2\% drop (by incorrectly penalizing the model for cached tool result mismatches), and engineering innovations (minimum token generation, pre-closed thinking, cache synchronization) collectively enable reliable training. The final system improves validation reward from 0.43 to 0.78 on Qwen3-4B-Thinking.

A central finding is that \textbf{RL provides training stability but limited improvement beyond a strong SFT baseline} in the cached rollout setting. Despite trying deterministic rewards (zero noise), increased model capacity (LoRA 128), larger datasets (44K samples), and multiple judge configurations, RL consistently plateaus within $\pm$0.01 of the SFT starting point. This suggests that the ceiling is set by the cache fidelity---with 50\% of rollouts receiving non-exact cached tool results, the reward signal quality fundamentally limits what RL can learn.

\paragraph{Limitations.}
The cached rollout approach is inherently limited by cache coverage: tool calls with novel arguments receive degraded responses, introducing systematic noise. Our reliance on GPT-5 as a judge introduces both cost and potential inconsistency. The pre-closed thinking approach causes 16\% meta-reasoning leakage that we mitigate but cannot eliminate.

\paragraph{Future Work.}
The most impactful next step is transitioning from cached to \emph{live tool execution}, which would eliminate the cache fidelity bottleneck and enable genuine RL improvement. Other directions include scaling to larger models (7B--14B) where RL may have more room to improve, training a dedicated reward model to replace the noisy LLM-as-judge, and extending to multi-agent collaborative tool-use settings.